\documentclass{article}

\usepackage{times}
\usepackage{graphicx} 
\usepackage{subfigure} 

\usepackage{natbib}

\usepackage{algorithm}
\usepackage{algorithmic}

\usepackage{hyperref}

\usepackage[accepted]{icml2015}

\icmltitlerunning{Deep Unsupervised Learning using Nonequilibrium Thermodynamics}

\usepackage{amssymb}
\usepackage{pdflscape}
\usepackage[export]{adjustbox}
\graphicspath{ {figures/} }
\usepackage{amsmath}

\newcommand{\pd}[2]{\frac{\partial #1}{\partial #2}}

\newcommand{\mb}{\mathbf}
\newcommand{\mc}{\mathcal}

\newcommand{\argmax}{\operatornamewithlimits{argmax}}

\newcommand{\pf}{q\left( \mb x^{(t)} | \mb x^{(t-1)} \right)}

\newcommand{\pr}{q\left( \mb x^{(t-1)} | \mb x^{(t)} \right)}

\newcommand{\qr}{p\left( \mb x^{(t-1)} | \mb x^{(t)} \right)}
\newcommand{\qrtil}{\tilde{p}\left( \mb x^{(t-1)} | \mb x^{(t)} \right)}
\newcommand{\qrhat}{\hat{p}\left( \mb x^{(t-1)} | \mb x^{(t)} \right)}
\newcommand{\pst}{q \left( \mb x^{(0)} \right)}

\newcommand{\qst}{p \left( \mb x^{(T)} \right)}
\newcommand{\ptraj}{q\left( \mb x^{(0\cdots T)} \right)}

\newcommand{\qtrajtil}{\tilde{p}\left( \mb x^{(0\cdots T)} \right)}
\newcommand{\pcondtraj}{q\left( \mb x^{(1\cdots T)} | \mb x^{(0)} \right)}
\newcommand{\qtraj}{p\left( \mb x^{(0\cdots T)} \right)}
\newcommand{\ptarget}{\pi\left( \mb x^{(T)} \right)}
\newcommand{\qmarg}{p\left( \mb x^{(0)} \right)}
\newcommand{\pmarg}{q\left( \mb x^{(T)} \right)}

\usepackage{tablefootnote}

\begin{document}

\twocolumn[

\icmltitle{Deep Unsupervised Learning using\\
Nonequilibrium Thermodynamics
}

\icmlauthor{Jascha Sohl-Dickstein}{jascha@stanford.edu}
\icmladdress{Stanford University}

\icmlauthor{Eric A. Weiss}{eaweiss@berkeley.edu}
\icmladdress{University of California, Berkeley}

\icmlauthor{Niru Maheswaranathan}{nirum@stanford.edu}
\icmladdress{Stanford University}

\icmlauthor{Surya Ganguli}{sganguli@stanford.edu}
\icmladdress{Stanford University}

\icmlkeywords{generative models, nonequilibrium thermodynamics, unsupervised learning}

\vskip 0.3in

]

\begin{abstract}
A central problem in machine learning involves modeling complex data-sets using highly flexible families of probability distributions in which learning, sampling, inference, and evaluation are still analytically or computationally tractable. Here, we develop an approach that simultaneously achieves both flexibility and tractability. The essential idea, inspired by non-equilibrium statistical physics, is to systematically and slowly destroy structure in a data distribution through an iterative forward diffusion process. We then learn a reverse diffusion process that restores structure in data, yielding a highly flexible and tractable generative model of the data. This approach allows us to rapidly learn, sample from, and evaluate probabilities in deep generative models with thousands of layers or time steps, as well as to compute conditional and posterior probabilities under the learned model. We additionally release an open source reference implementation of the algorithm.
\end{abstract}

\section{Introduction}

Historically, probabilistic models suffer from a tradeoff between two conflicting objectives: \textit{tractability} and \textit{flexibility}. Models that are \textit{tractable} can be analytically evaluated and easily fit to data (e.g. a Gaussian or Laplace). However, these models are unable to aptly describe structure in rich datasets. On the other hand, models that are \textit{flexible} can be molded to fit structure in arbitrary data. For example, we can define models in terms of any (non-negative) function $\phi(\mb x)$ yielding the flexible distribution 
$p\left(\mb x\right) = \frac{\phi\left(\mb x
\right)}{Z}$, where $Z$ is a normalization constant. However, computing this normalization constant is generally intractable. Evaluating, training, or drawing samples from such flexible models typically requires a very 
expensive Monte Carlo process.

A variety of analytic approximations exist which ameliorate, but do not remove, this tradeoff--for instance mean field theory and
its  
expansions \cite{plefka,Tanaka:1998p1984}, variational Bayes \cite{jordan1999introduction}, 
contrastive divergence \cite{Welling:2002p3,Hinton02}, 
minimum probability flow \cite{MPF_ICML,SohlDickstein2011a}, minimum KL contraction \cite{mkc}, 
proper scoring rules \cite{gneiting2007strictly,parry2012proper}, 
score matching \cite{Hyvarinen05}, pseudolikelihood \cite{besag}, loopy belief propagation \cite{murphy1999loopy}, 
and many, many more.
Non-parametric methods \cite{gershman2012tutorial} 
can also be very effective\footnote{Non-parametric methods can be seen as transitioning smoothly between tractable and flexible models.  For instance, a non-parametric Gaussian mixture model will represent a small amount of data using a single Gaussian, 
but may represent infinite data as a mixture of an infinite number of Gaussians.}.

\subsection{Diffusion probabilistic models}
We present a novel way to define probabilistic models that allows:\\[-2.2em]
\begin{enumerate}\itemsep1pt \parskip0pt \parsep0pt
  \item extreme flexibility in model structure,
  \item exact sampling,
  \item easy multiplication with other distributions, e.g. in order to compute a posterior, and
  \item the model log likelihood, and the probability of individual states, to be cheaply evaluated.
\end{enumerate}
~\\[-2.3em]
Our method uses a Markov chain to gradually convert one distribution into another, an idea used in non-equilibrium statistical physics \cite{Jarzynski:1997p12846} and sequential Monte Carlo \cite{Neal:AIS}. We build a generative Markov chain which converts a simple known distribution (e.g. a Gaussian) into a target 
(data) distribution using a diffusion process. Rather than use this Markov chain to approximately evaluate a model which 
has been otherwise defined, we explicitly define the probabilistic model as the endpoint of the Markov chain. Since each step in the diffusion 
chain has an analytically evaluable probability, the full chain can also be analytically evaluated.

Learning in this framework involves estimating small perturbations to a diffusion process. Estimating small 
perturbations is 
more tractable than explicitly describing the full distribution with a single, non-analytically-normalizable, potential function.  Furthermore, since a diffusion 
process exists for any smooth target distribution, this method can capture data distributions of arbitrary form.

We demonstrate the utility of these \textit{diffusion probabilistic models} by training high log likelihood models for a two-dimensional swiss roll, binary sequence, handwritten digit (MNIST), and several natural image (CIFAR-10, bark, and dead leaves) datasets.

\subsection{Relationship to other work}
\begin{figure*}
\centering
\begin{tabular}{cccc}
& $t=0$ & $t=\frac{T}{2}$ & $t=T$
\\
$\ptraj$
 &
\raisebox{-.43\height}{
\includegraphics[width=0.18\linewidth]{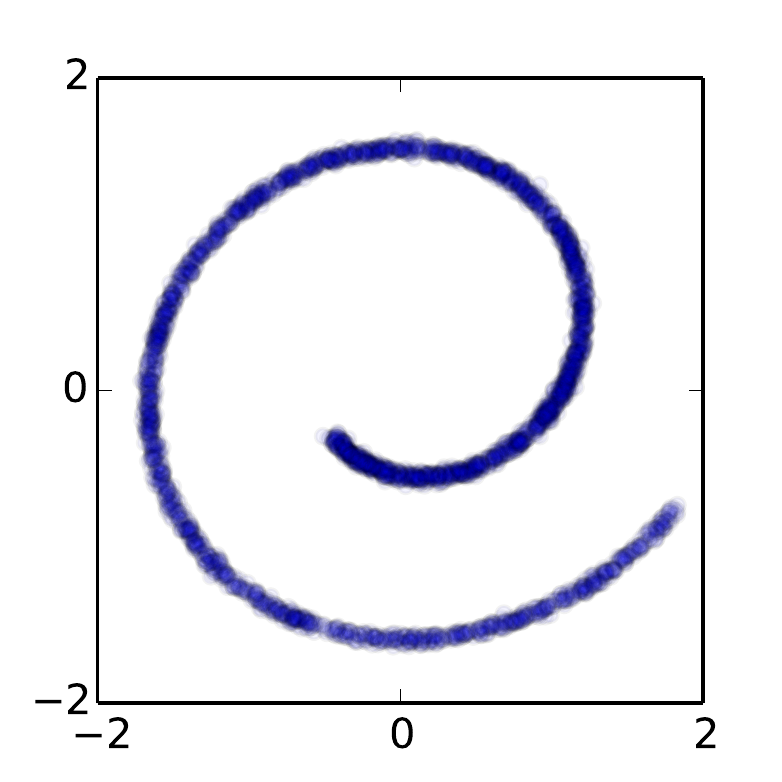} 
}
&
\raisebox{-.43\height}{
\includegraphics[width=0.18\linewidth]{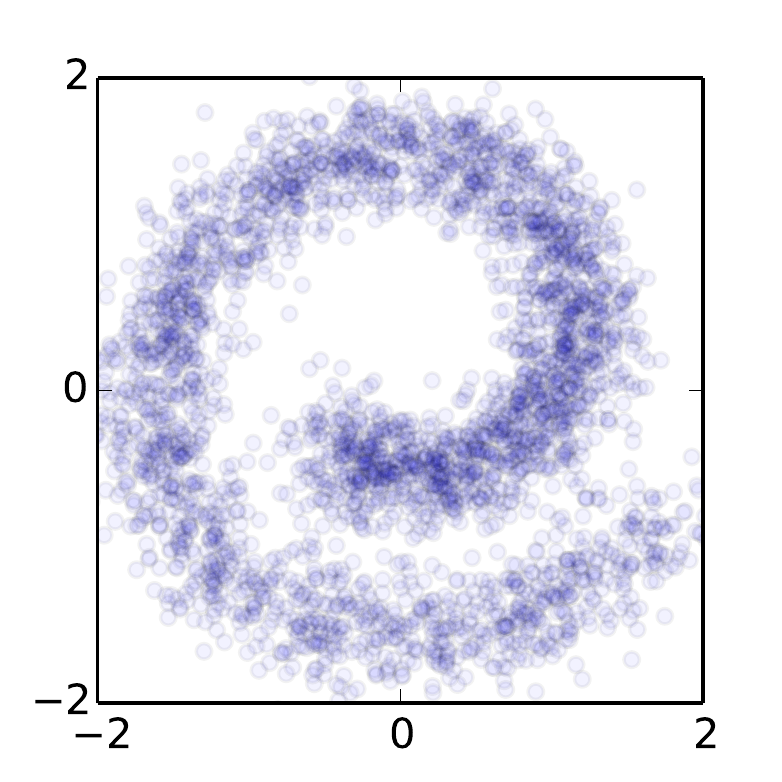} 
}
&
\raisebox{-.43\height}{
\includegraphics[width=0.18\linewidth]{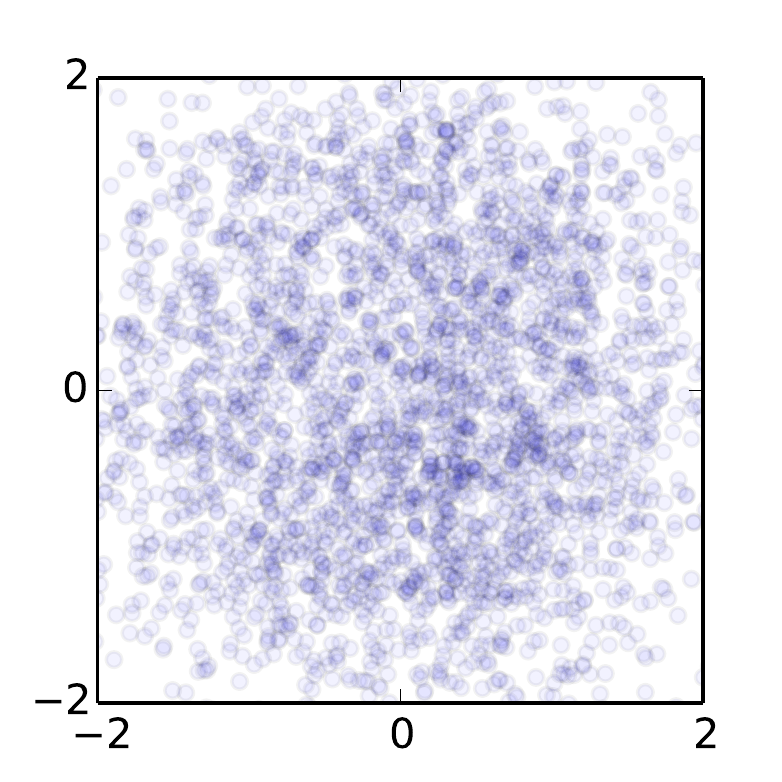} 
}
\\
$\qtraj$ &
\raisebox{-.43\height}{
\includegraphics[width=0.18\linewidth]{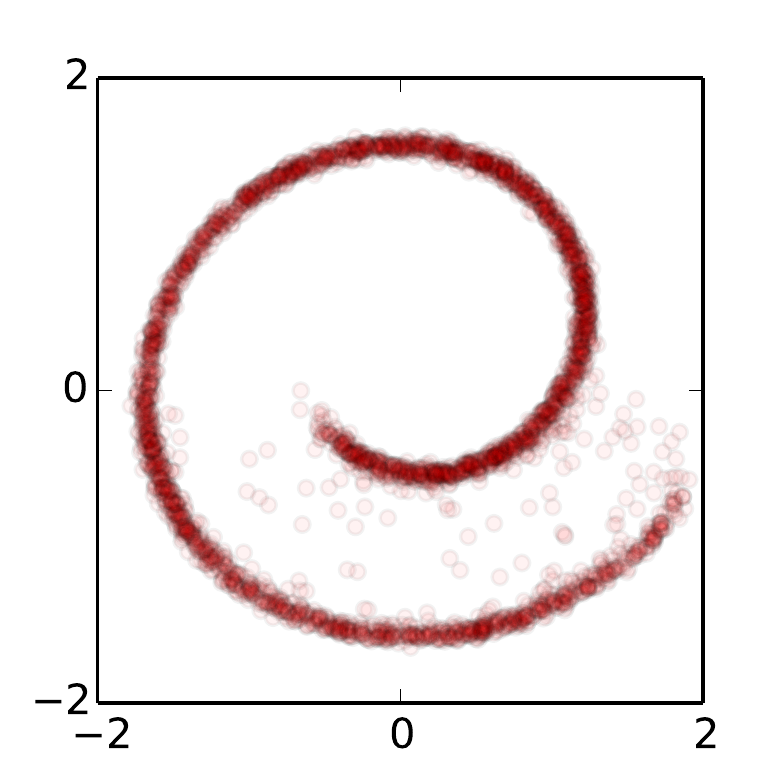} 
}
&
\raisebox{-.43\height}{
\includegraphics[width=0.18\linewidth]{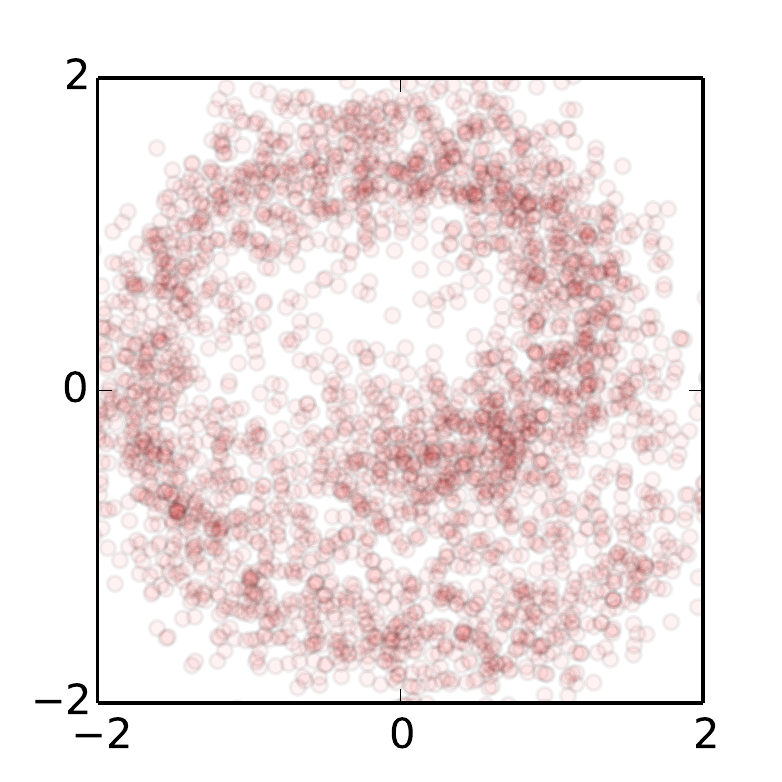} 
}
&
\raisebox{-.43\height}{
\includegraphics[width=0.18\linewidth]{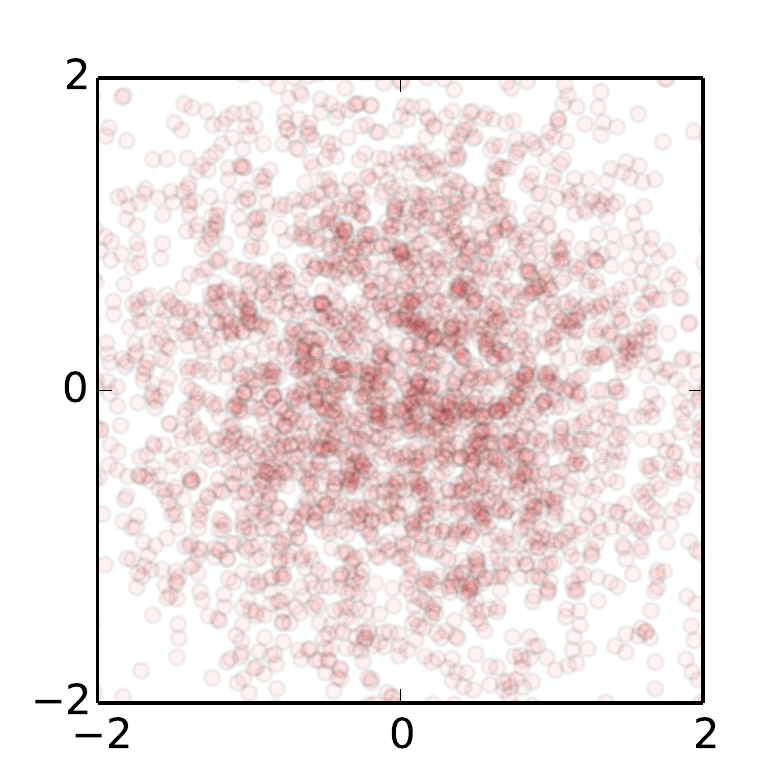} 
}
\\
$\mb f_\mu\left( \mb x^{(t)}, t \right) - \mb x^{(t)}$ &
\raisebox{-.43\height}{
\includegraphics[width=0.18\linewidth]{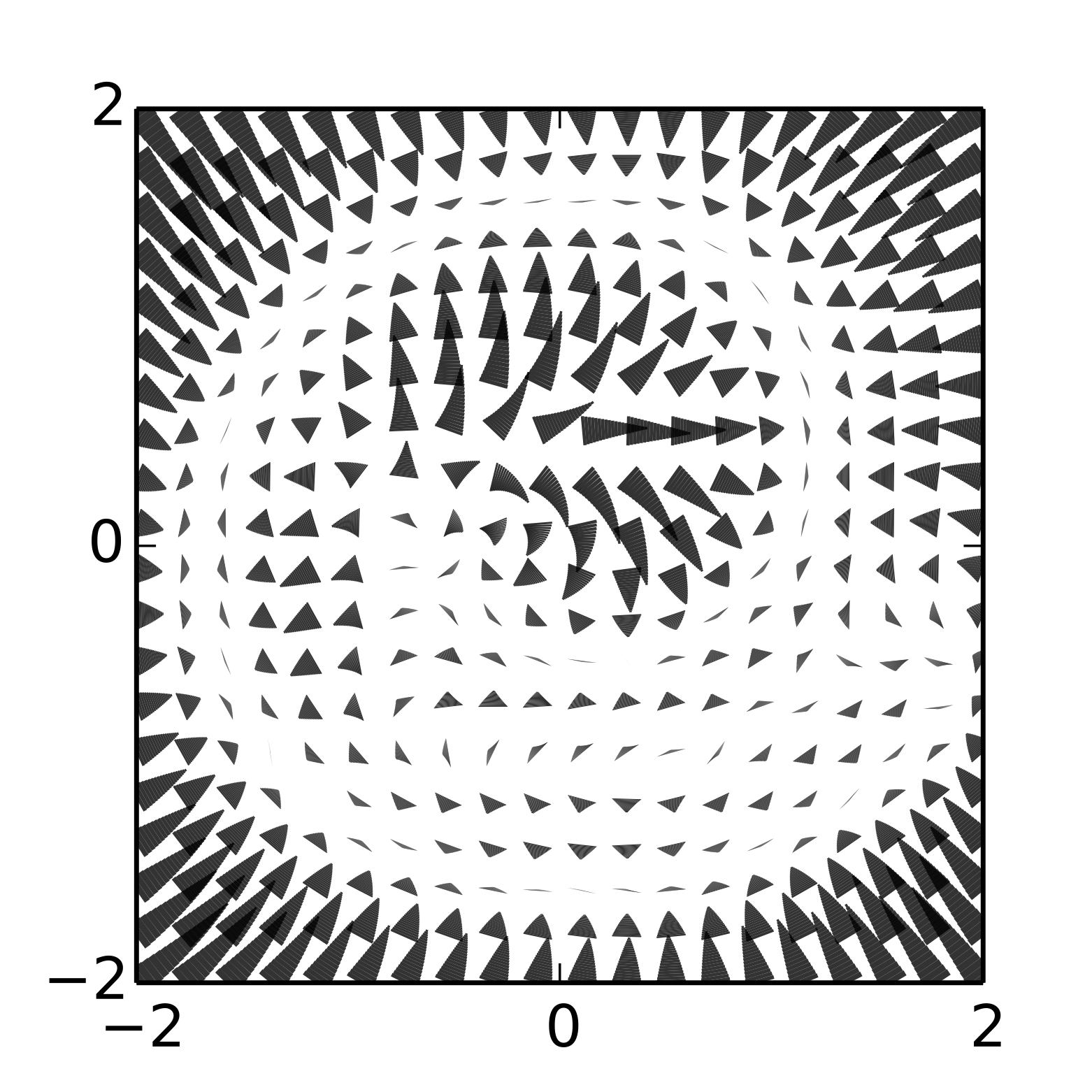} 
}
&
\raisebox{-.43\height}{
\includegraphics[width=0.18\linewidth]{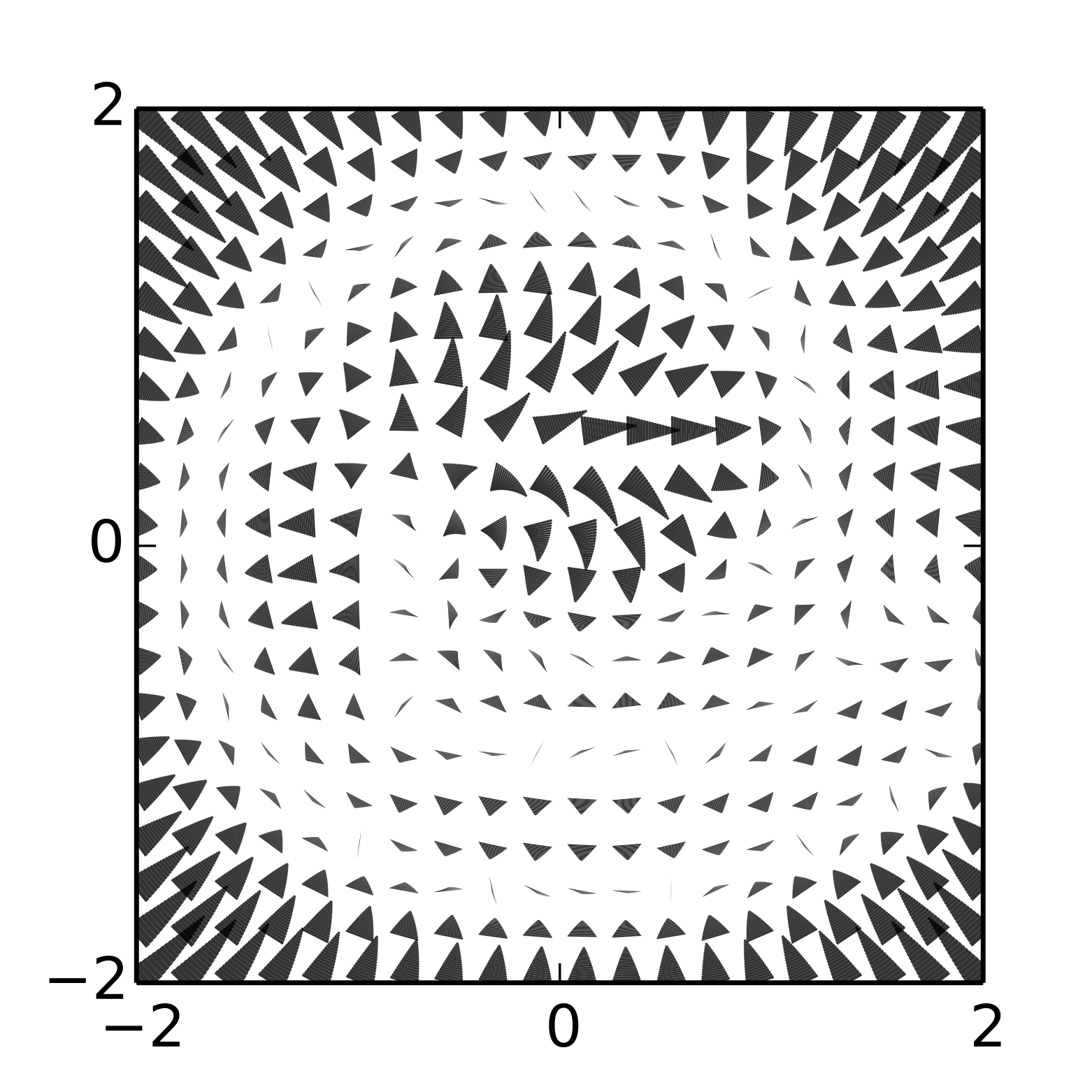} 
}
&
\raisebox{-.43\height}{
\includegraphics[width=0.18\linewidth]{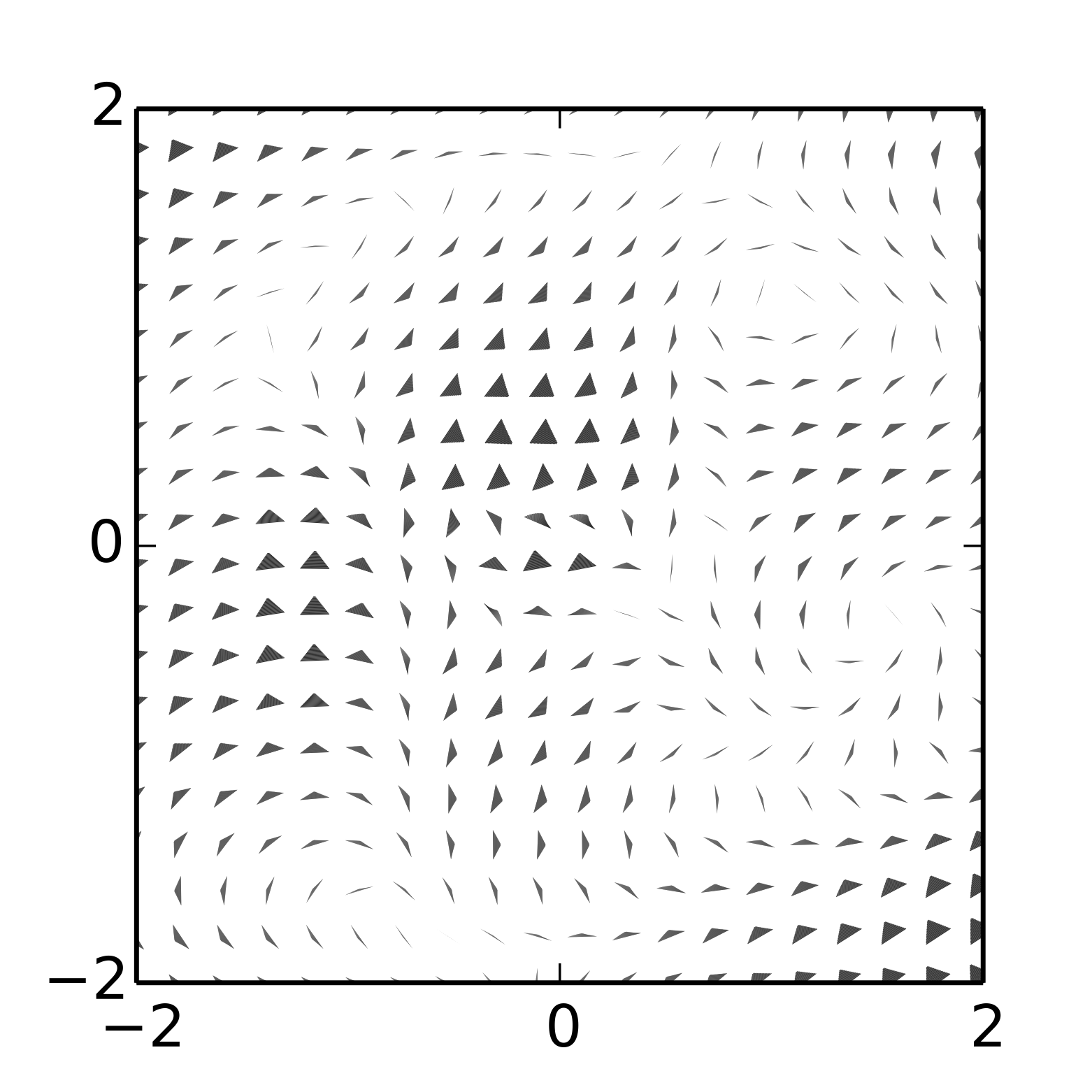} 
}
\end{tabular}
\caption{
The proposed modeling framework trained on 2-d swiss roll data.  The top row shows time slices from the forward trajectory $\ptraj$.  The data 
distribution (left) undergoes Gaussian diffusion, which gradually transforms it into an identity-covariance Gaussian (right).  The middle row shows 
the corresponding time slices from the trained reverse trajectory $\qtraj$.  An identity-covariance Gaussian (right) undergoes a Gaussian diffusion 
process with learned mean and covariance functions, and is gradually transformed back into the data distribution (left).
The bottom row shows the drift term, $\mb f_\mu\left( \mb x^{(t)}, t \right) - \mb x^{(t)}$, for the same reverse diffusion process.
}
\label{fig swiss}
\end{figure*}

The wake-sleep algorithm \cite{Hinton1995,dayan1995helmholtz} introduced the idea of training inference and generative probabilistic models against each other.  This approach remained largely unexplored for nearly two decades, though with some exceptions \cite{sminchisescu2006learning,kavukcuoglu2010fast}.
There has been a recent explosion of work 
developing this idea. In 
\cite{Kingma2013,Gregor2013,Rezende2014,Ozair2014} 
variational learning and inference algorithms were developed which allow a 
flexible generative model and posterior distribution over latent variables to be directly trained against each other. 

The variational bound in these papers 
is similar to the one used in our training objective and in the earlier work of \cite{sminchisescu2006learning}.
However, our motivation and model form are both quite different, and the present work 
retains the following differences and advantages relative to these techniques:\\[-2.0em]
\begin{enumerate}\itemsep1pt \parskip0pt \parsep0pt
	\item We develop our framework using ideas from physics, quasi-static processes, and annealed importance sampling rather than  
from variational Bayesian methods.
	\item We show how to easily multiply the learned distribution with another probability distribution (eg with a conditional distribution in order to 
compute a posterior)
	\item We address the difficulty that training the inference model can prove particularly challenging in variational inference methods, due to the asymmetry in the objective between the inference and generative models. We restrict the forward (inference) 
process to a simple functional form, in such a way that the reverse (generative) process will have the same functional form.
	\item We train models with thousands of layers (or time steps), rather than only a handful of layers.
	\item We provide upper and lower bounds on the entropy production in each layer (or time step)
\end{enumerate}
~\\[-1.7em]
There are a number of related techniques for training probabilistic models (summarized below) that develop highly flexible forms for generative models,
train stochastic trajectories, or learn the reversal of a Bayesian network.
Reweighted wake-sleep \cite{Bornschein2014} develops extensions and improved learning rules for the original wake-sleep algorithm.
Generative stochastic networks \cite{bengio2013deep,yao2014equivalence} 
train a Markov kernel to match its equilibrium distribution to the data distribution. 
Neural autoregressive distribution estimators \cite{Larochelle2011} (and their recurrent \cite{Uria2013} and deep \cite{Uria2013a} extensions) decompose 
a joint distribution into a sequence of tractable conditional distributions over each dimension. 
Adversarial networks \cite{goodfellowgenerative} train a generative model against a classifier which attempts to 
distinguish generated samples from true data. 
A similar objective in \cite{Schmidhuber1992} learns a two-way mapping to a representation with marginally independent units.
In \cite{Rippel2013,Dinh2014} bijective deterministic maps are learned to a latent representation with a simple factorial density function.
In \cite{Stuhlmuller2013} stochastic inverses are learned for Bayesian networks.
Mixtures of conditional Gaussian scale mixtures (MCGSMs) \cite{theis2012mixtures} 
describe a dataset using Gaussian scale mixtures, with parameters 
which depend on a sequence of causal neighborhoods. 
There is additionally significant work learning flexible generative mappings from simple latent distributions to 
data distributions -- early examples including \cite{MacKay1995} where neural networks are introduced as generative models, and 
\cite{Bishop1998} where a stochastic manifold mapping is learned from a latent space to the data space. 
We will compare experimentally against adversarial networks and MCGSMs.

Related ideas from physics include the Jarzynski equality \cite{Jarzynski:1997p12846}, known in machine learning as Annealed Importance Sampling (AIS) \cite{Neal:AIS},
which uses a 
Markov chain which slowly converts one distribution into another to compute a 
ratio of normalizing constants. 
In \cite{Burda2014} it is shown that AIS can also be performed using the reverse rather than forward trajectory.
Langevin dynamics \cite{langevin1908theorie}, which are the stochastic realization of the Fokker-Planck equation, show how 
to define a 
Gaussian diffusion process which has any target distribution as its equilibrium. In \cite{Suykens1995} the Fokker-Planck equation is used to perform stochastic optimization. 
Finally, the Kolmogorov forward and backward equations \cite{feller1949theory} show that for many forward diffusion processes, 
the reverse diffusion processes can be described using the same functional form.

%

\begin{figure*}
\centering
\begin{tabular}{cccc}
& $t=0$ & $t=\frac{T}{2}$ & $t=T$
\\
$\qtraj$ &
\raisebox{-.52\height}{
\includegraphics[width=0.175\linewidth]{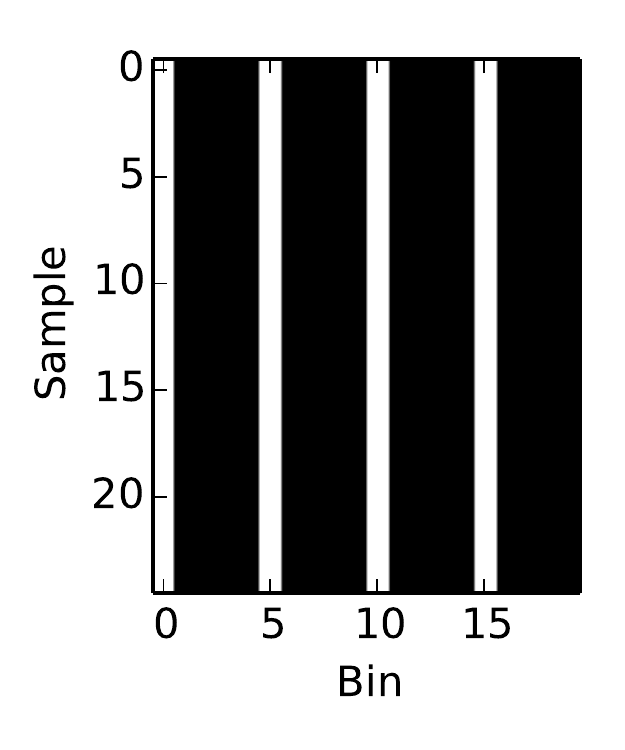} 
}
&
\raisebox{-.52\height}{
\includegraphics[width=0.175\linewidth]{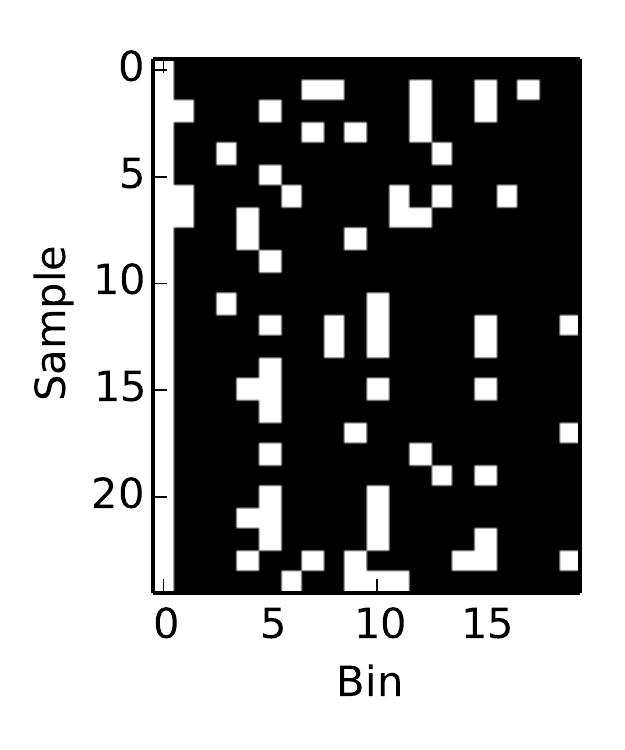} 
}
&
\raisebox{-.52\height}{
\includegraphics[width=0.175\linewidth]{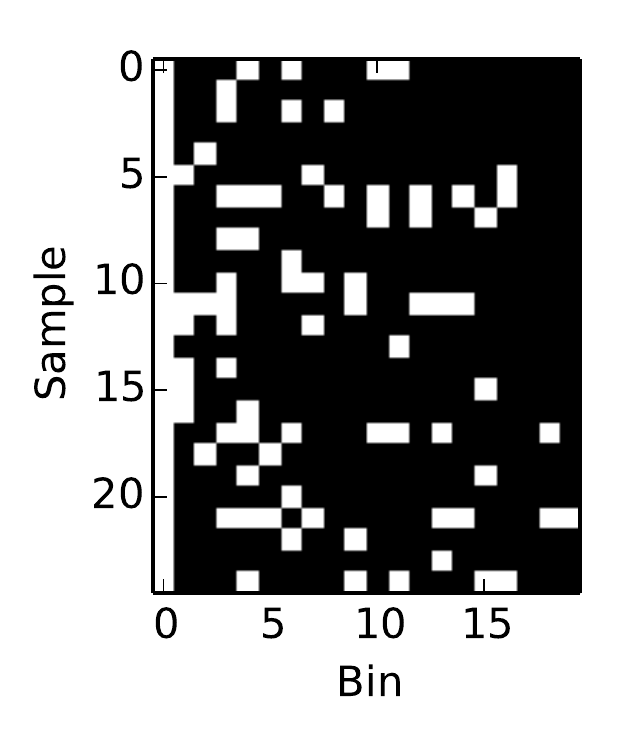} 
}
\end{tabular}
\caption{
Binary sequence learning via binomial diffusion. A binomial diffusion model was trained on binary `heartbeat' data, where a pulse occurs every 5th bin. Generated samples (left) are identical to the training data.  The sampling procedure consists of initialization at independent binomial noise (right), which is then transformed into the data distribution by a binomial diffusion process, with trained bit flip probabilities. Each row contains an independent sample. For ease of visualization, all samples have been shifted so that a pulse occurs in the first column. In the raw sequence data, the first pulse is uniformly distributed over the first five bins.
}
\label{fig heartbeat}
\end{figure*}

\begin{figure*}
\centering
\begin{tabular}{cc}
(a)\adjincludegraphics[width=0.3\linewidth,trim={{0.23\width} {0.2\width} {0.1\width} {0.1\width}},clip]{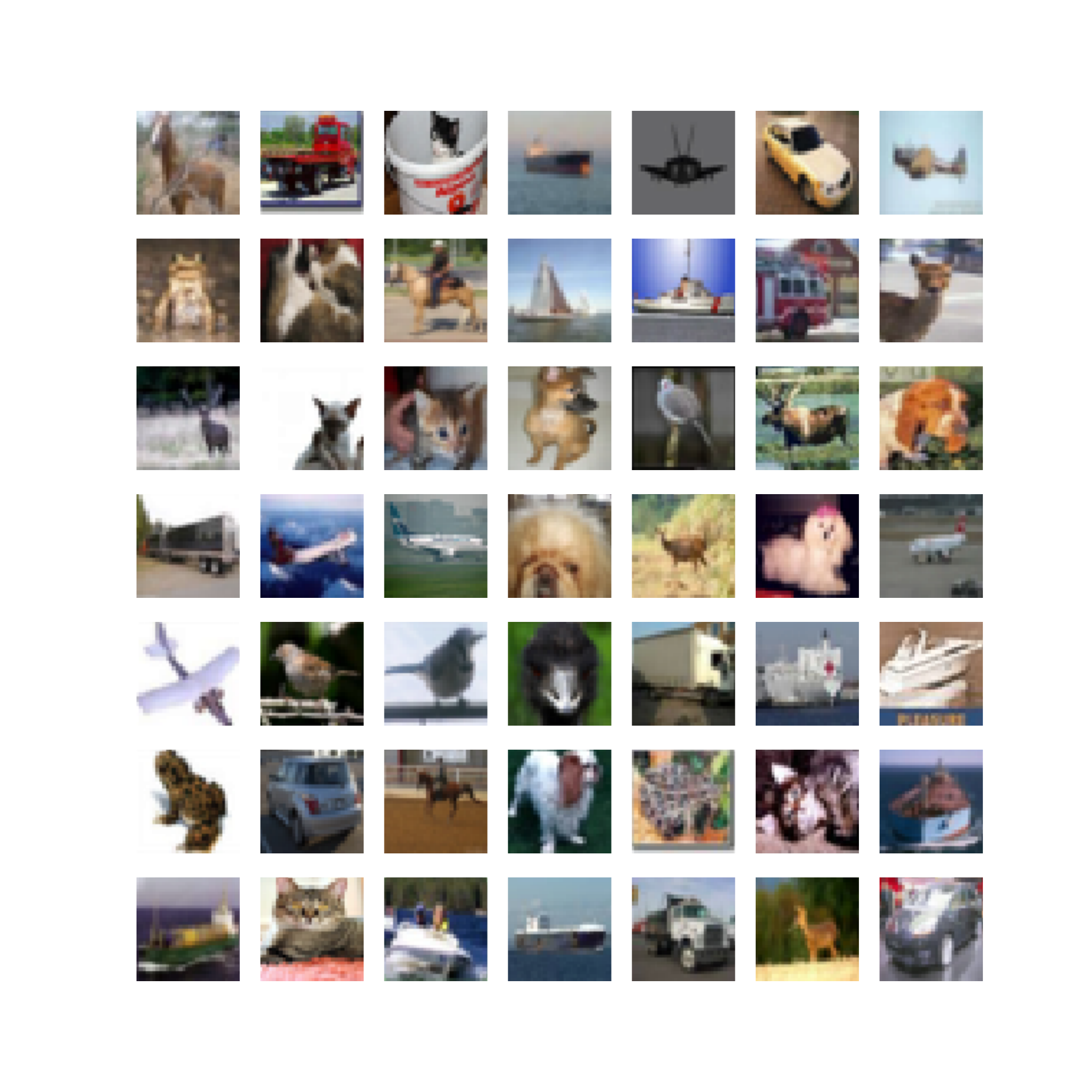} &
(b)\adjincludegraphics[width=0.3\linewidth,trim={{0.23\width} {0.2\width} {0.1\width} {0.1\width}},clip]{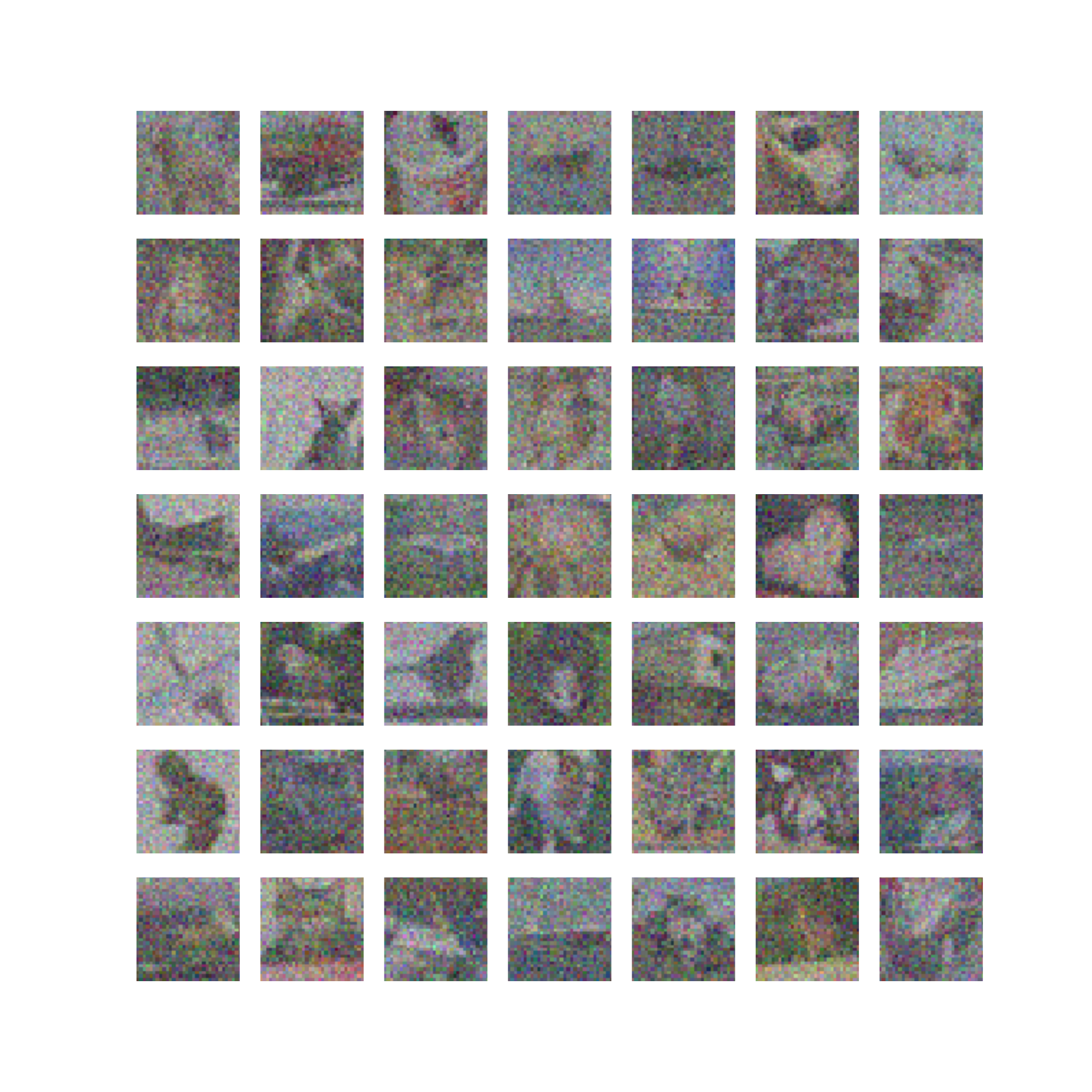}  \\
(c)\adjincludegraphics[width=0.3\linewidth,trim={{0.23\width} {0.2\width} {0.1\width} {0.1\width}},clip]{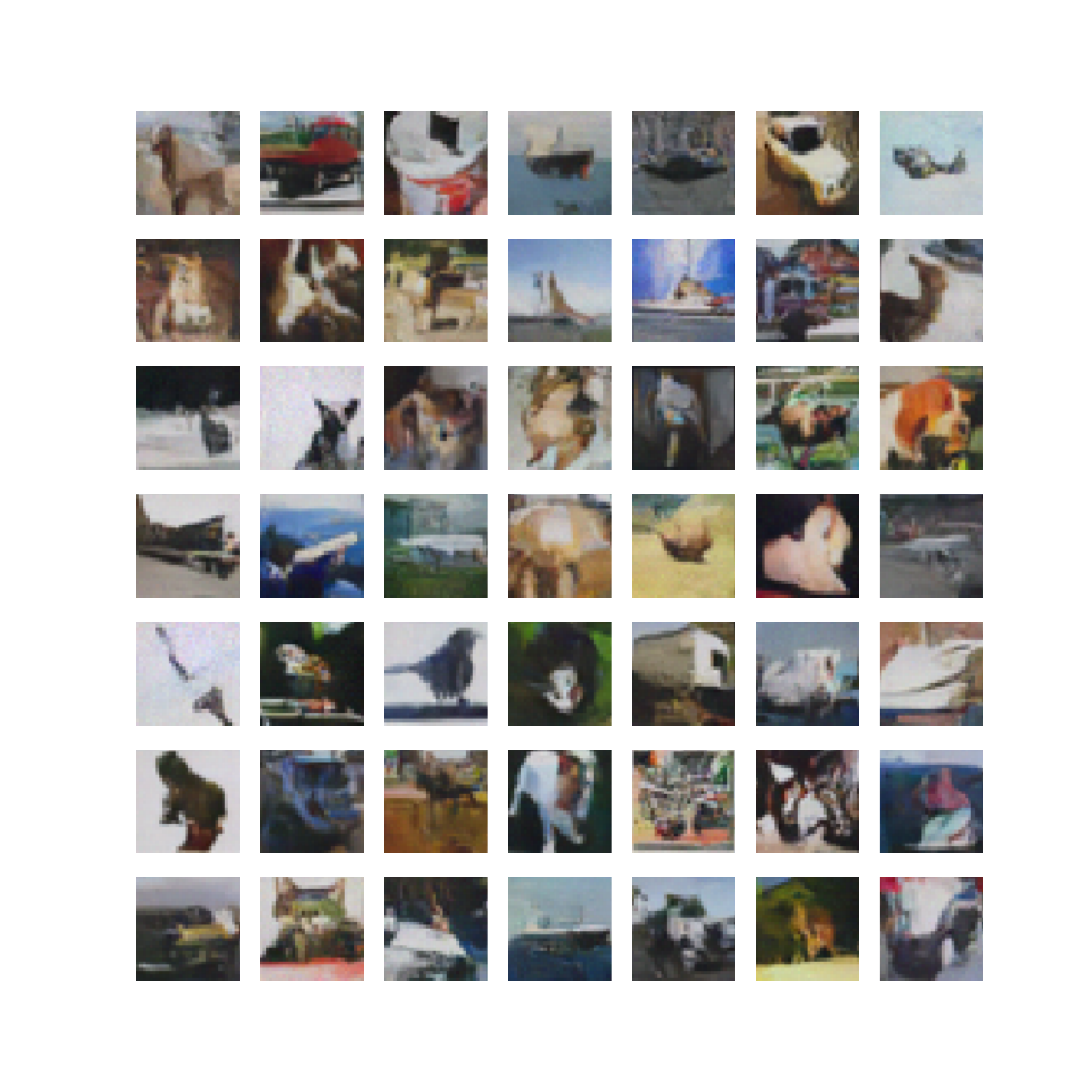} &
(d)\adjincludegraphics[width=0.3\linewidth,trim={{0.23\width} {0.2\width} {0.1\width} {0.1\width}},clip]{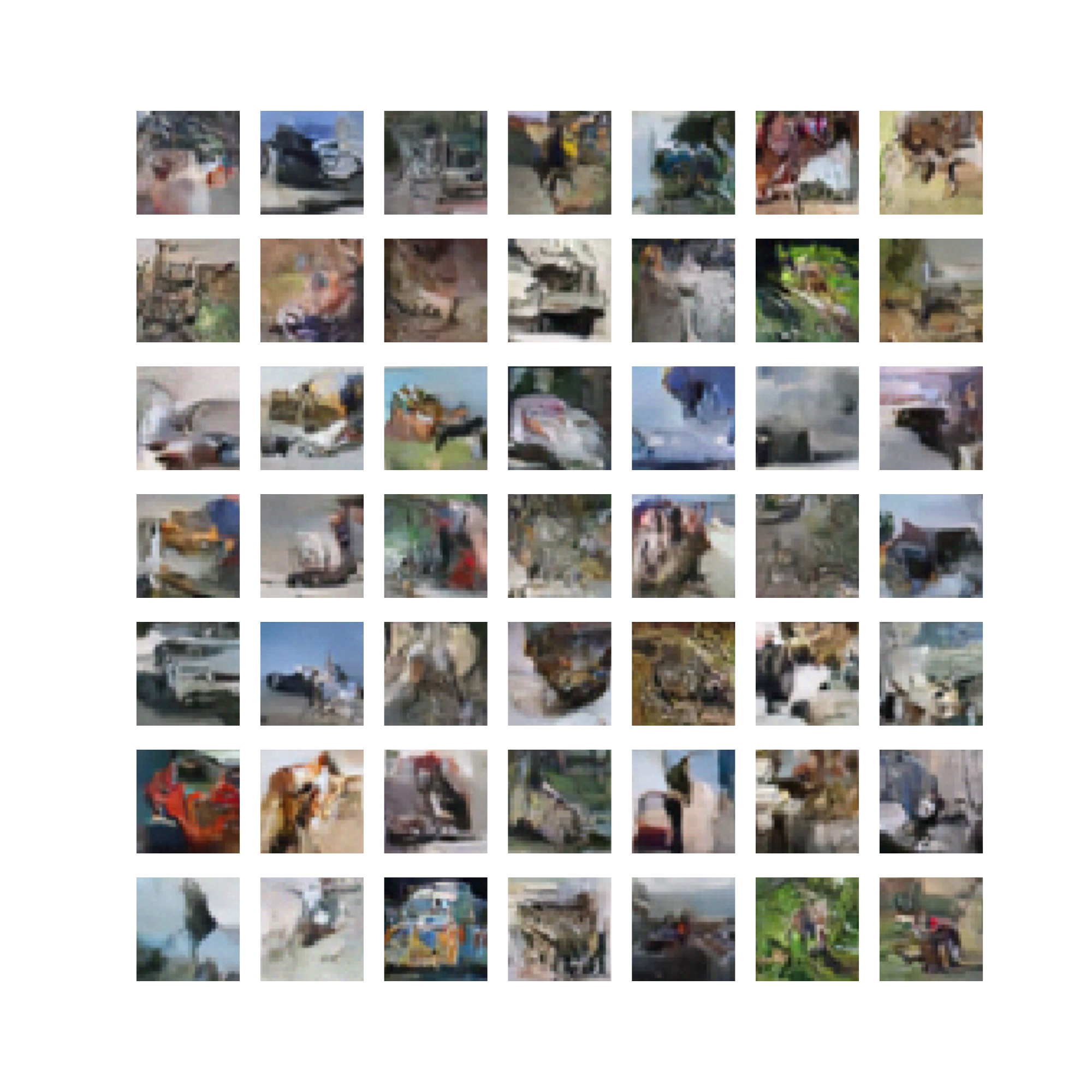} 
\end{tabular}
\caption{
The proposed framework trained on the CIFAR-10 \cite{Krizhevsky2009} dataset. 
{\em (a)} Example holdout data (similar to training data).
{\em (b)} Holdout data corrupted with Gaussian noise of variance 1 (SNR = 1). 
{\em (c)} Denoised images, generated by sampling from the posterior distribution over denoised images conditioned on the images in (b).
{\em (d)} Samples generated by the diffusion model.
}
\label{fig cifar}
\end{figure*}

\section{Algorithm}

Our goal is to define a forward (or inference) diffusion process which converts any complex data distribution into a simple, tractable, distribution, and then learn a finite-time reversal of this diffusion process which defines our generative model distribution (See Figure \ref{fig swiss}). 
We first describe the forward, inference diffusion process.  
We then show how the reverse, generative diffusion process can be trained and used to evaluate probabilities.  
We also derive entropy bounds for the reverse process, and show how the learned distributions can be multiplied by any second distribution 
(e.g. as would be done to compute a posterior when inpainting or denoising an image). 

\subsection{Forward Trajectory}
\label{sec forward}

We label the data distribution $\pst$.  The data distribution is gradually converted into a well behaved (analytically tractable) distribution 
$\pi\left( \mb y \right)$ by repeated application of a Markov diffusion kernel $T_\pi\left( \mb y | \mb y'; \beta \right)$ for $\pi\left( \mb y \right)$, 
where $\beta$ is the diffusion rate,
\begin{align}
\pi\left( \mb y \right) &= \int d\mb y' T_\pi\left( \mb y | \mb y'; \beta \right) \pi\left( \mb y' \right) \\
\pf &= T_\pi\left( \mb x^{(t)} | \mb x^{(t-1)}; \beta_t \right)
.
\end{align}
The forward trajectory, corresponding to starting at the data distribution and performing $T$ steps of diffusion, is thus
\begin{align}
\ptraj &= \pst \prod_{t=1}^T \pf
\end{align}
For the experiments shown below, $\pf$ corresponds to either Gaussian diffusion into a Gaussian distribution with identity-covariance, or binomial 
diffusion into an independent binomial distribution.
Table \ref{tab diff} gives the diffusion kernels for both Gaussian and binomial distributions.  

\subsection{Reverse Trajectory}
\label{sec reverse}
The generative distribution will be trained to describe the same trajectory, but in reverse,
\begin{align}
\qst &= \ptarget \\
\qtraj &= \qst \prod_{t=1}^T \qr
.
\end{align}
For both Gaussian and binomial diffusion, for continuous diffusion (limit of small step size $\beta$) the reversal of the diffusion process has the 
identical functional form as the forward process \cite{feller1949theory}. 
Since $\pf$ is a Gaussian (binomial) distribution, and if $\beta_t$ is small, then $\pr$ will also be a Gaussian (binomial) 
distribution.  The longer the trajectory the smaller the diffusion rate $\beta$ can be made.

During learning only the mean and covariance for a Gaussian diffusion kernel, or the bit flip probability for a binomial kernel, need be estimated.  As 
shown in Table \ref{tab diff}, $\mb f_\mu\left( \mb x^{(t)}, t \right)$ and $\mb f_\Sigma\left( \mb x^{(t)}, t \right)$ are functions defining the mean and 
covariance of the reverse Markov transitions for a Gaussian, and $\mb f_b\left( \mb x^{(t)}, t \right)$ is a function providing the bit flip probability for 
a binomial distribution. 
The computational cost of running this algorithm is the cost of these functions, times the number of time-steps. 
For all results in this paper, multi-layer perceptrons are used to define these functions.  A wide range of regression or function fitting techniques 
would be applicable however, including nonparameteric methods.

\subsection{Model Probability}
\label{sec mod prob}
The probability the generative model assigns to the data is
\begin{align}
p\left( \mb x^{(0)} \right) &= \int d\mb x^{(1\cdots T)} \qtraj
.
\end{align}
Naively this integral is intractable -- but taking a cue from annealed importance sampling and the Jarzynski equality, we instead evaluate the relative 
probability of the forward and reverse trajectories, averaged over forward trajectories,
\begin{align}
\qmarg &= \int d\mb x^{(1\cdots T)} \qtraj \frac{\pcondtraj}{\pcondtraj} \\
&= \int d\mb x^{(1\cdots T)} \pcondtraj \frac{\qtraj}{\pcondtraj} \\
&= \int d\mb x^{(1\cdots T)} \pcondtraj \cdot \nonumber \\ &\qquad \quad \qst \prod_{t=1}^T \frac{\qr}{\pf}
.
\end{align}
This can be evaluated rapidly by averaging over samples from the forward trajectory $\pcondtraj$.  For 
infinitesimal $\beta$ the forward and reverse distribution over trajectories can be made identical (see Section \ref{sec reverse}).  
If they are identical then only a {\em single} 
sample from $\pcondtraj$ is required to exactly evaluate the above integral, as can be seen by substitution.  This corresponds to the case of a 
quasi-static process in statistical 
physics \cite{spinney2013fluctuation,jarzynski2013equalities}.

\subsection{Training}
Training amounts to maximizing the model log likelihood,\\[-1.5em]
\begin{align}
L &= \int d\mb x^{(0)} \pst \log \qmarg \\
&= \int d\mb x^{(0)} \pst \cdot \nonumber \\ & \qquad \log \left[
	\begin{array}{l}
		 \int d\mb x^{(1\cdots T)} \pcondtraj \cdot \\
		\qquad \qst \prod_{t=1}^T \frac{\qr}{\pf}
	\end{array}
	 \right] \label{eq pre jen},
\end{align}
which has a lower bound provided by Jensen's inequality,
\begin{align}
L &\geq \int d\mb x^{(0 \cdots T)} \ptraj \cdot \nonumber \\ & \qquad \log \left[ \qst \prod_{t=1}^T \frac{\qr}{\pf} \right] \label{eq post jen}
.
\end{align}
As described in Appendix \ref{app bound}, for our diffusion trajectories this reduces to,
\begin{align}
L
&\geq K \label{eq ineq LK}\\
K = & 
-\sum_{t=2}^T \int d\mb x^{(0)}d\mb x^{(t)} q\left( \mb x^{(0)}, \mb x^{(t)} \right) \cdot
\nonumber\\  & \qquad \qquad
	D_{KL}\left( 
		q\left( \mb x^{(t-1)} | \mb x^{(t)}, \mb x^{(0)} \right)
			\middle|\middle|
		\qr
	\right)  \nonumber \\
&	+ H_q\left( \mb X^{(T)} | \mb X^{(0)} \right) - H_q\left( \mb X^{(1)} | \mb X^{(0)} \right)
	- H_p\left( \mb X^{(T)} \right)
.
\end{align}
where the entropies and KL divergences can be analytically computed. The derivation of this bound 
parallels the derivation of the log likelihood bound in variational Bayesian methods.

As in Section \ref{sec mod prob} if the forward and reverse 
trajectories are identical, corresponding to a quasi-static process, then the inequality in Equation \ref{eq ineq LK} becomes an equality.

Training consists of finding the reverse Markov transitions which maximize this lower bound on the log likelihood,
\begin{align}
\qrhat &= \argmax_{\qr}
	K 
.
\end{align}
The specific targets of estimation for Gaussian and binomial diffusion are given in 
Table \ref{tab diff}.

Thus, the task of estimating a probability distribution has been reduced to the task of 
performing regression on the functions which set the mean and covariance of a sequence
of Gaussians (or set the state flip probability for a sequence of Bernoulli trials).

\subsubsection{Setting the Diffusion Rate $\beta_t$}

The choice of $\beta_t$ in the forward trajectory is important for the performance of the trained model.  In AIS, the right schedule of intermediate 
distributions can greatly improve the accuracy of the log partition function estimate \cite{grosse2013annealing}. 
In thermodynamics the schedule taken when moving 
between equilibrium distributions determines how much free energy is lost \cite{spinney2013fluctuation,jarzynski2013equalities}. 

In the case of Gaussian diffusion, we learn\footnote{Recent experiments suggest that it is just as effective to instead use the same fixed $\beta_t$ schedule as for binomial diffusion.} the forward diffusion schedule $\beta_{2\cdots T}$ by gradient ascent on $K$. The variance $\beta_1$ of the first step is fixed to a small constant to prevent overfitting.  The dependence of samples from $\pcondtraj$ on $\beta_{1\cdots T}$ is made explicit by using `frozen noise' -- as in \cite{Kingma2013} the noise is treated as an additional auxiliary variable, and held constant while computing partial derivatives of $K$ with respect to the parameters.

For binomial diffusion, the discrete state space makes gradient ascent with frozen noise impossible. 
We instead choose the forward diffusion schedule $\beta_{1\cdots T}$ to erase a constant fraction $\frac{1}{T}$ of the original signal per diffusion step, yielding a diffusion rate of $\beta_t = \left( T-t +1\right)^{-1}$.

\begin{figure*}
\centering
\begin{tabular}{ccc}
(a)\adjincludegraphics[width=0.28\linewidth,trim={{0.16\width} {0.05\width} {0.17\width} {0.06\width}},clip]{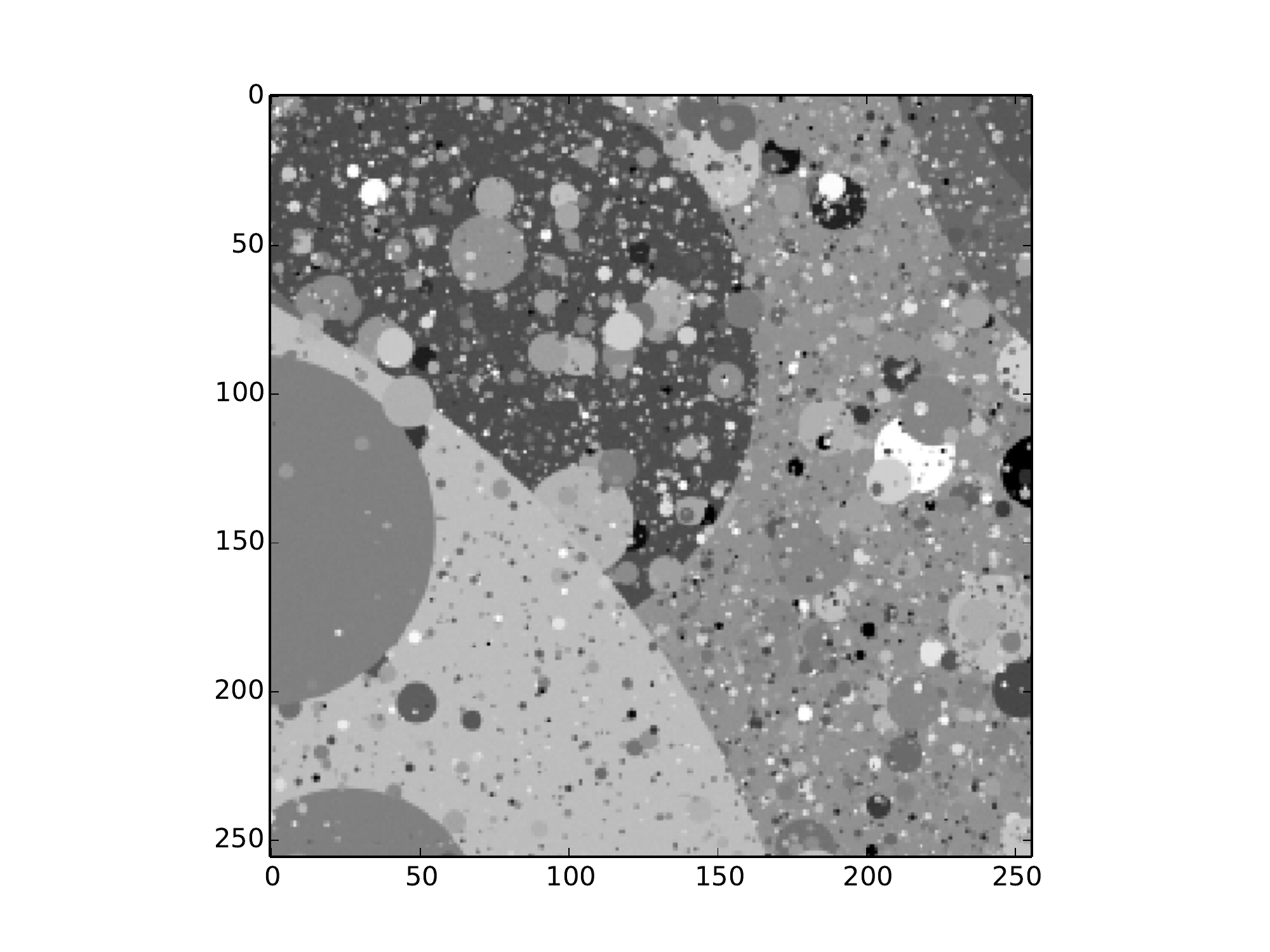} &
(b)\adjincludegraphics[width=0.28\linewidth,trim={{0.16\width} {0.05\width} {0.17\width} {0.06\width}},clip]{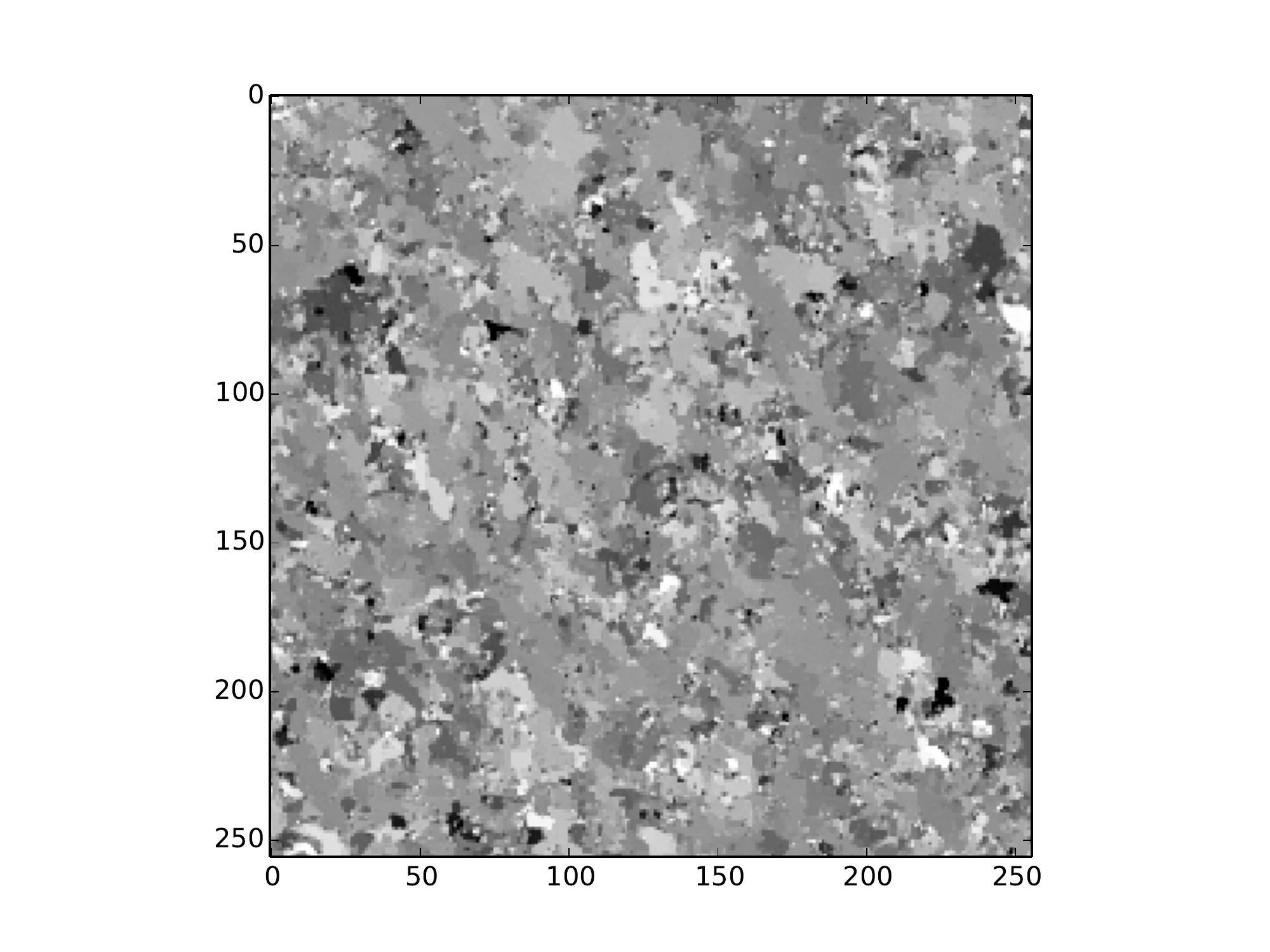} &
(c)\adjincludegraphics[width=0.28\linewidth,trim={{0.16\width} {0.05\width} {0.17\width} {0.06\width}},clip]{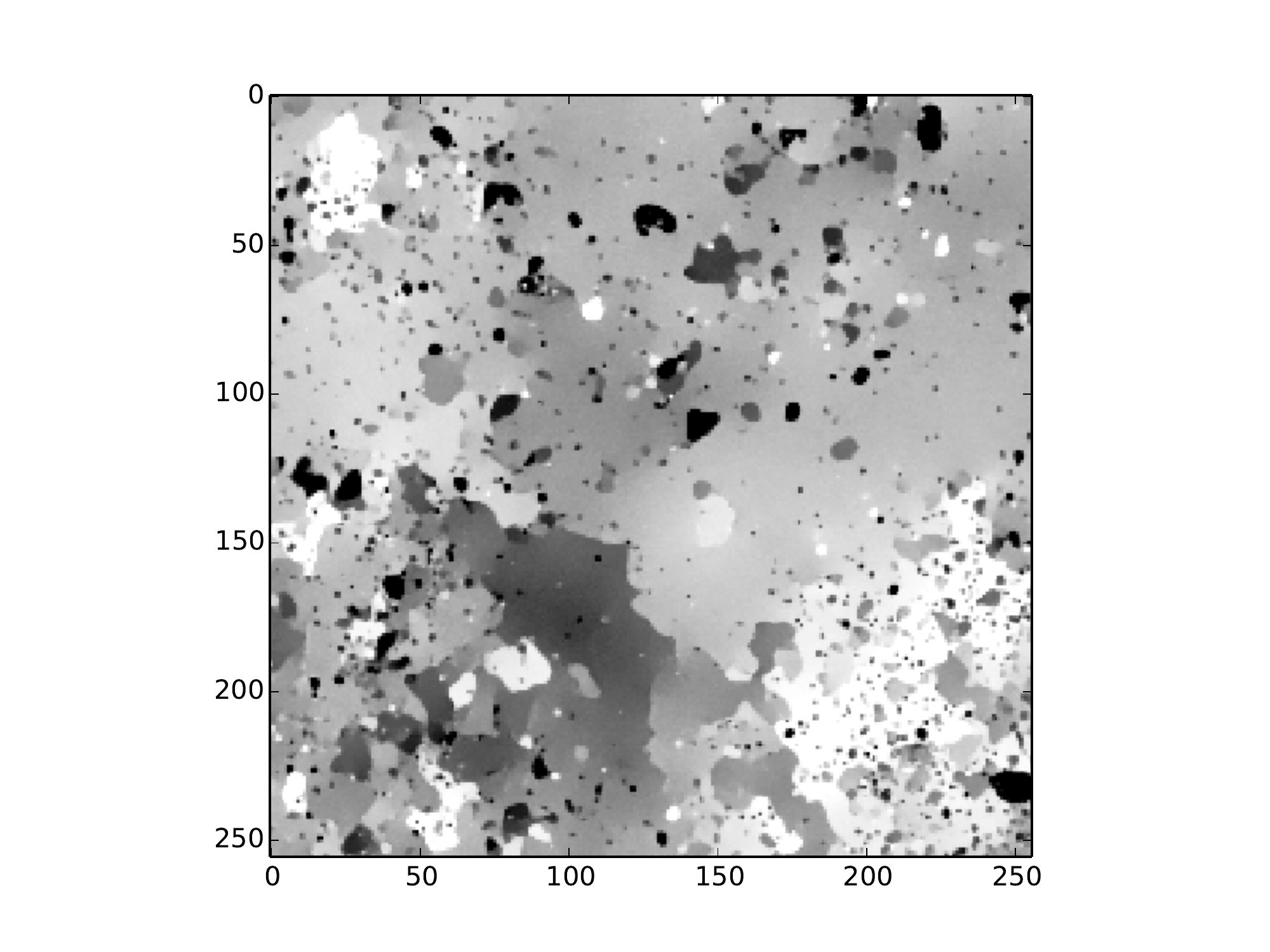}
\end{tabular}
\caption{
The proposed framework trained on dead leaf images \cite{Jeulin1997,Lee2001}. 
{\em (a)} Example training image. 
{\em (b)} A sample from the previous state of the art natural image model \cite{theis2012mixtures} trained on identical data, reproduced here with permission. 
{\em (c)} A sample generated by the diffusion model. Note that it demonstrates fairly consistent occlusion relationships, displays a 
multiscale distribution over object sizes, and produces circle-like objects, especially at smaller scales. 
As shown in Table \ref{tb ll compare}, the diffusion model has the highest log likelihood on the test set.
}
\label{fig dead leaf}
\end{figure*}

\subsection{Multiplying Distributions, and Computing Posteriors}
\label{sec posterior}

Tasks such as computing a posterior in order to do signal 
denoising or inference of missing values requires multiplication of the model distribution $\qmarg$ with a second distribution, or bounded positive function, $r\left(\mb x^{(0)} \right)$, 
producing a new distribution $\tilde{p}\left(\mb x^{(0)} \right) \propto \qmarg r\left(\mb x^{(0)} \right)$.

Multiplying distributions is costly and difficult for many techniques, including variational autoencoders, GSNs, NADEs, 
and most graphical models. However, under a diffusion model it is straightforward, since the second distribution can be treated 
either as a small perturbation to each step in the diffusion process, or often exactly multiplied into each diffusion step. 
Figures \ref{fig cifar} and \ref{fig bark} demonstrate the use of a diffusion model to perform denoising and inpainting of 
natural images. 
The following sections describe how to multiply distributions in the context of diffusion probabilistic models.


\subsubsection{Modified Marginal Distributions}

First, in order to compute $\tilde{p}\left(\mb x^{(0)} \right)$, we multiply each of the intermediate distributions by a corresponding function $r
\left( \mb x^{(t)} \right)$. We use a tilde above a distribution or Markov transition to denote that it belongs to a trajectory that has been modified in this way.
$\qtrajtil$ is the modified reverse trajectory, which starts at the distribution $\tilde{p}\left( \mb x^{(T)} \right) = \frac{1}{\tilde{Z}_T} p\left( \mb x^{(T)} \right) r\left( \mb x^{(T)} \right)$ and proceeds through 
the sequence of intermediate distributions 
\begin{align}
\tilde{p} \left( \mb x^{(t)} \right) 
	& = 
\frac{1}{\tilde{Z}_t} p\left(\mb x^{(t)} \right) r\left( \mb x^{(t)} \right)
\label{eq perturb marginals}
,
\end{align}
where $\tilde{Z}_t$ is the normalizing constant for the $t$th intermediate distribution.

\subsubsection{Modified Diffusion Steps}

The Markov kernel $p\left( \mb x^{(t)} \mid \mb x^{(t+1)}\right)$ for the reverse diffusion process obeys the equilibrium condition
\begin{align}
p\left( \mb x^{(t} \right) &= \int d \mb x^{(t+1)} p\left( \mb x^{t)} \mid \mb x^{(t+1)} \right) p\left( \mb x^{t+1)} \right)
.
\end{align}
We wish the perturbed Markov kernel $\tilde{p}\left( \mb x^{(t)} \mid \mb x^{(t+1)}\right)$ to instead obey the equilibrium condition for the perturbed distribution,
\begin{align}
\tilde{p}\left( \mb x^{(t)} \right) &= \int d \mb x^{(t+1)} \tilde{p}\left( \mb x^{(t)} \mid \mb x^{(t+1)} \right) \tilde{p}\left( \mb x^{t+1)} \right), \\
\frac{
	p\left(\mb x^{(t)} \right) r\left( \mb x^{(t)} \right)
	}{\tilde{Z}_t}
&= \int d \mb x^{(t+1)} \tilde{p}\left( \mb x^{(t)} \mid \mb x^{(t+1)} \right) 
\cdot \nonumber\\  & \qquad \qquad
	\frac{
	p\left(\mb x^{(t+1)} \right) r\left( \mb x^{(t+1)} \right)
	}{\tilde{Z}_{t+1}}
,\\
p\left(\mb x^{(t)} \right)
&= \int d \mb x^{(t+1)} \tilde{p}\left( \mb x^{(t)} \mid \mb x^{(t+1)} \right) 
\cdot \nonumber\\  & \qquad \qquad
	\frac{
	 	\tilde{Z}_t r\left( \mb x^{(t+1)} \right)
	}{
		\tilde{Z}_{t+1} r\left( \mb x^{(t)} \right)
	} p\left(\mb x^{(t+1)} \right)
	\label{eq tild fixed point}
.
\end{align}

Equation \ref{eq tild fixed point} will be satisfied if
\begin{align}
\tilde{p}\left( \mb x^{(t)} | \mb x^{(t+1)} \right) &= p\left( \mb x^{(t)} | \mb x^{(t+1)} \right) 
	\frac{
		\tilde{Z}_{t+1} r\left( \mb x^{(t)} \right)
	}{
	 	\tilde{Z}_t r\left( \mb x^{(t+1)} \right)
	}
\label{eq tild unnorm}
.
\end{align}
Equation \ref{eq tild unnorm} may not correspond to a normalized probability distribution, so we choose $\tilde{p}\left( \mb x^{(t)} | \mb x^{(t+1)} \right)$ to be the corresponding normalized distribution
\begin{align}
\tilde{p}\left( \mb x^{(t)} | \mb x^{(t+1)} \right) &= \frac{1}{\tilde{Z}_t\left( \mb x^{(t+1)} \right)} p\left( \mb x^{(t)} | \mb x^{(t+1)} \right) r\left( \mb x^{(t)} \right)
\label{eq tild norm}
,
\end{align}
where $\tilde{Z}_t\left( \mb x^{(t+1)} \right)$ is the normalization constant.

For a Gaussian, each diffusion step is typically very sharply peaked relative to $r\left( \mb x^{(t)} \right)$, due to its small variance.
This means that $\frac{
		r\left( \mb x^{(t)} \right)
	}{
	 	r\left( \mb x^{(t+1)} \right)
	}$
can be treated as a small perturbation to $p\left( \mb x^{(t)} | \mb x^{(t+1)} \right)$.
A small perturbation to a Gaussian effects the mean, but not the normalization constant,
so in this case Equations \ref{eq tild unnorm} and \ref{eq tild norm} are equivalent
(see Appendix \ref{sec perturb derive}).

\subsubsection{Applying $r\left( \mb x^{(t)} \right)$}

If $r\left( \mb x^{(t)} \right)$ is sufficiently smooth, 
then it can be treated as a small
perturbation to the reverse diffusion kernel $p\left( \mb x^{(t)} | \mb x^{(t+1)} \right)$. In this case $\tilde{p}\left( \mb x^{(t)} | \mb x^{(t+1)} \right)$ will have an identical functional form to $p\left( \mb x^{(t)} | \mb x^{(t+1)} \right)$, 
but with perturbed mean for the Gaussian kernel, or with perturbed flip rate for the binomial kernel. 
The perturbed diffusion kernels 
are given in Table \ref{tab diff}, and are derived for the Gaussian in Appendix \ref{sec perturb derive}.

If $r\left( \mb x^{(t)} \right)$ can be multiplied with a Gaussian (or binomial) distribution in closed form,
then it can be directly multiplied with the reverse diffusion kernel $p\left( \mb x^{(t)} | \mb x^{(t+1)} \right)$ in closed form.
This applies in the case where $r\left( \mb x^{(t)} \right)$ consists of a delta function for some subset of coordinates, as in
the inpainting example in Figure \ref{fig bark}.

\subsubsection{Choosing $r\left( \mb x^{(t)} \right)$}

Typically, $r\left( \mb x^{(t)} \right)$ should be chosen to change slowly over the course of the trajectory.  For the experiments in this paper we chose it to be constant,
\begin{align}
r\left( \mb x^{(t)} \right) &= r\left( \mb x^{(0)} \right).
\end{align}
Another convenient choice is $r\left( \mb x^{(t)} \right) = r\left( \mb x^{(0)} \right)^\frac{T-t}{T}$.  
Under this second choice $r\left( \mb x^{(t)} \right)$ makes no contribution to the starting distribution for the reverse trajectory. 
This guarantees that drawing the initial sample from 
$\tilde{p}\left( \mb x^{(T)} \right)$ for the reverse trajectory remains straightforward.

\subsection{Entropy of Reverse Process}

Since the forward process is known, we can derive upper and lower bounds on the conditional entropy of each step in the reverse trajectory, and thus on the log likelihood,
\begin{align}
H_q\left( \mb X^{(t)} | \mb X^{(t-1)} \right) + H_q\left( \mb X^{(t-1)} | \mb X^{(0)} \right)  - H_q\left( \mb X^{(t)} | \mb X^{(0)} \right) 
\nonumber \\ \qquad
 \le
H_q\left( \mb X^{(t-1)} | \mb X^{(t)} \right)
\le
H_q\left( \mb X^{(t)} | \mb X^{(t-1)} \right)
,
\end{align}
where both the upper and lower bounds depend only on $\pcondtraj$, and can be analytically computed.  
The derivation is provided in Appendix \ref{app entropy}.



%

\begin{table}
\hspace{-3mm}
\begin{center}
\begin{tabular}{l|l|l}
{\em Dataset} & {\em $K$} & {\em $K - L_{null}$} \\
\hline
Swiss Roll & 2.35 bits & 6.45 bits \\
Binary Heartbeat & -2.414 bits/seq. & 12.024 bits/seq. \\
Bark & -0.55 bits/pixel & 1.5 bits/pixel \\
Dead Leaves & 1.489 bits/pixel & 3.536 bits/pixel \\
CIFAR-10\tablefootnote{
An earlier version of this paper reported higher log likelihood bounds on CIFAR-10.
These were the result of the model learning the 8-bit quantization of pixel values in the CIFAR-10 dataset.
The log likelihood bounds reported here are instead for data that has been pre-processed by adding uniform noise to remove pixel quantization, as recommended in \cite{theis2015note}.
}
 & $5.4\pm0.2$ bits/pixel & $11.5\pm0.2$  bits/pixel \\
MNIST & \multicolumn{2}{c}{
	See table \ref{tb ll compare}
}
\end{tabular}
\caption{The lower bound $K$ on the log likelihood, computed on a holdout set, for each of the trained models.  
See Equation \ref{eq post jen}. The right column 
is the improvement relative to an isotropic Gaussian or independent binomial distribution. 
$L_{null}$ is the log likelihood of $\pi\left( \mb x^{(0)} \right)$.
All datasets except for Binary Heartbeat were scaled by a constant to give them variance 1 before computing log likelihood.
\label{tb K}
}
\end{center}
\end{table}
\begin{table}
\begin{center}
\begin{tabular}{l|l}
{\em Model} & {\em Log Likelihood} \\
\hline
	\textbf{\em Dead Leaves}
\\
\ \ \ \ MCGSM & 1.244 bits/pixel \\
\ \ \ \ \textbf{Diffusion} & \textbf{$\mathbf{1.489}$ bits/pixel}\\
\hline
	\textbf{\em MNIST}
\\
\ \ \ \ Stacked CAE & $174 \pm 2.3$ bits \\
\ \ \ \ DBN & $199 \pm 2.9$ bits \\
\ \ \ \ Deep GSN & $309 \pm 1.6$ bits \\
\ \ \ \ \textbf{Diffusion} & \textbf{$\mathbf{317 \pm 2.7}$ bits} \\
\ \ \ \ Adversarial net & $325 \pm 2.9$ bits \\
\ \ \ \ Perfect model & $349 \pm 3.3$ bits 
\end{tabular}
\caption{Log likelihood comparisons to other algorithms. Dead leaves images were evaluated using identical training and test data as in \cite{theis2012mixtures}. 
MNIST log likelihoods were estimated using the Parzen-window code from \cite{goodfellowgenerative}, 
with values given in bits,
and show 
that our performance is comparable to other recent techniques.
The perfect model entry was computed by applying the Parzen code to samples from the training data.
\label{tb ll compare}
}
\end{center}
\end{table}

\begin{figure*}
\centering
\begin{tabular}{ccc}
(a)\adjincludegraphics[width=0.24\linewidth,trim={{0.16\width} {0.05\width} {0.17\width} {0.06\width}},clip]{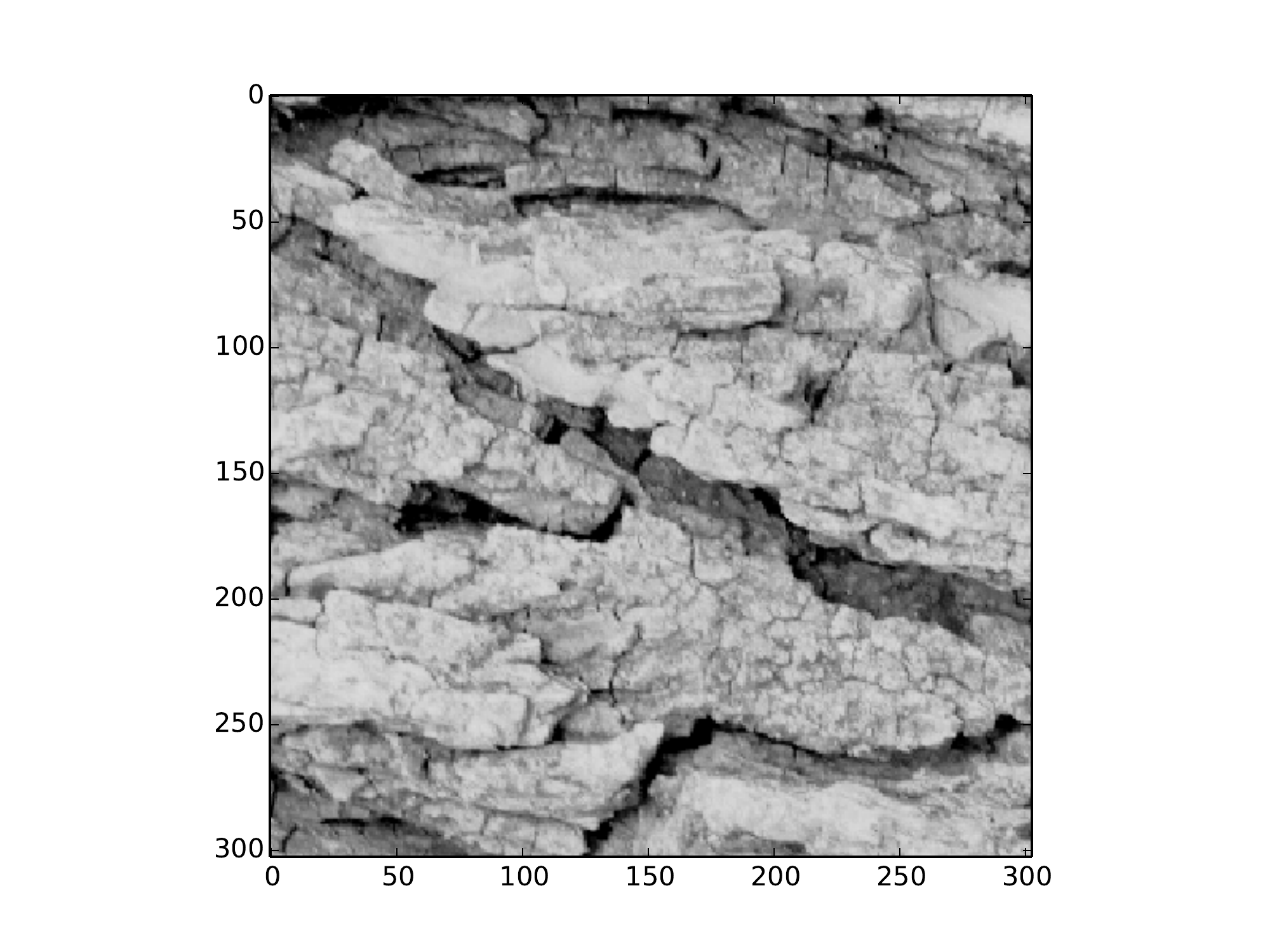} &
(b)\adjincludegraphics[width=0.24\linewidth,trim={{0.16\width} {0.05\width} {0.17\width} {0.06\width}},clip]{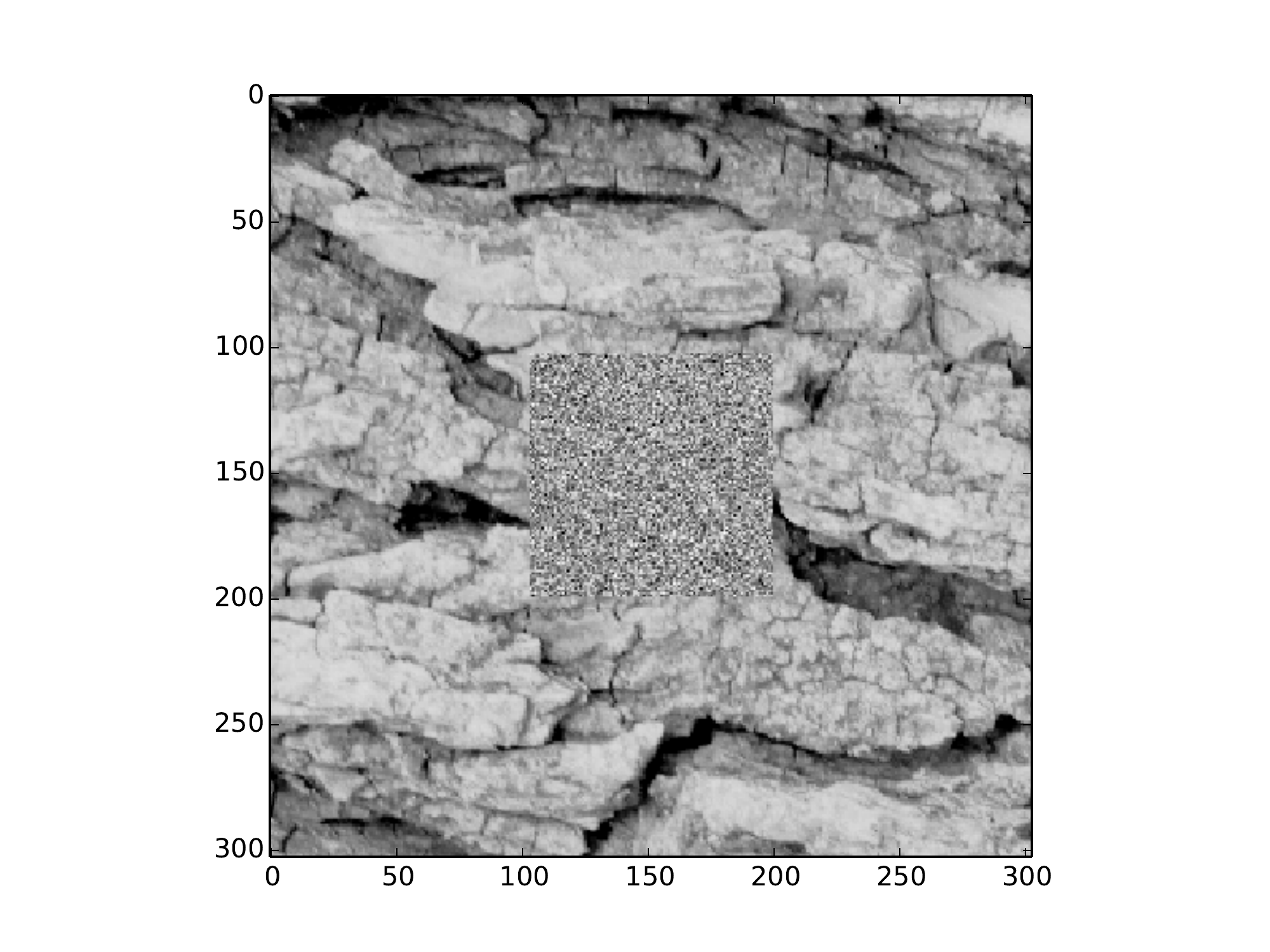} &
(c)\adjincludegraphics[width=0.24\linewidth,trim={{0.16\width} {0.05\width} {0.17\width} {0.06\width}},clip]{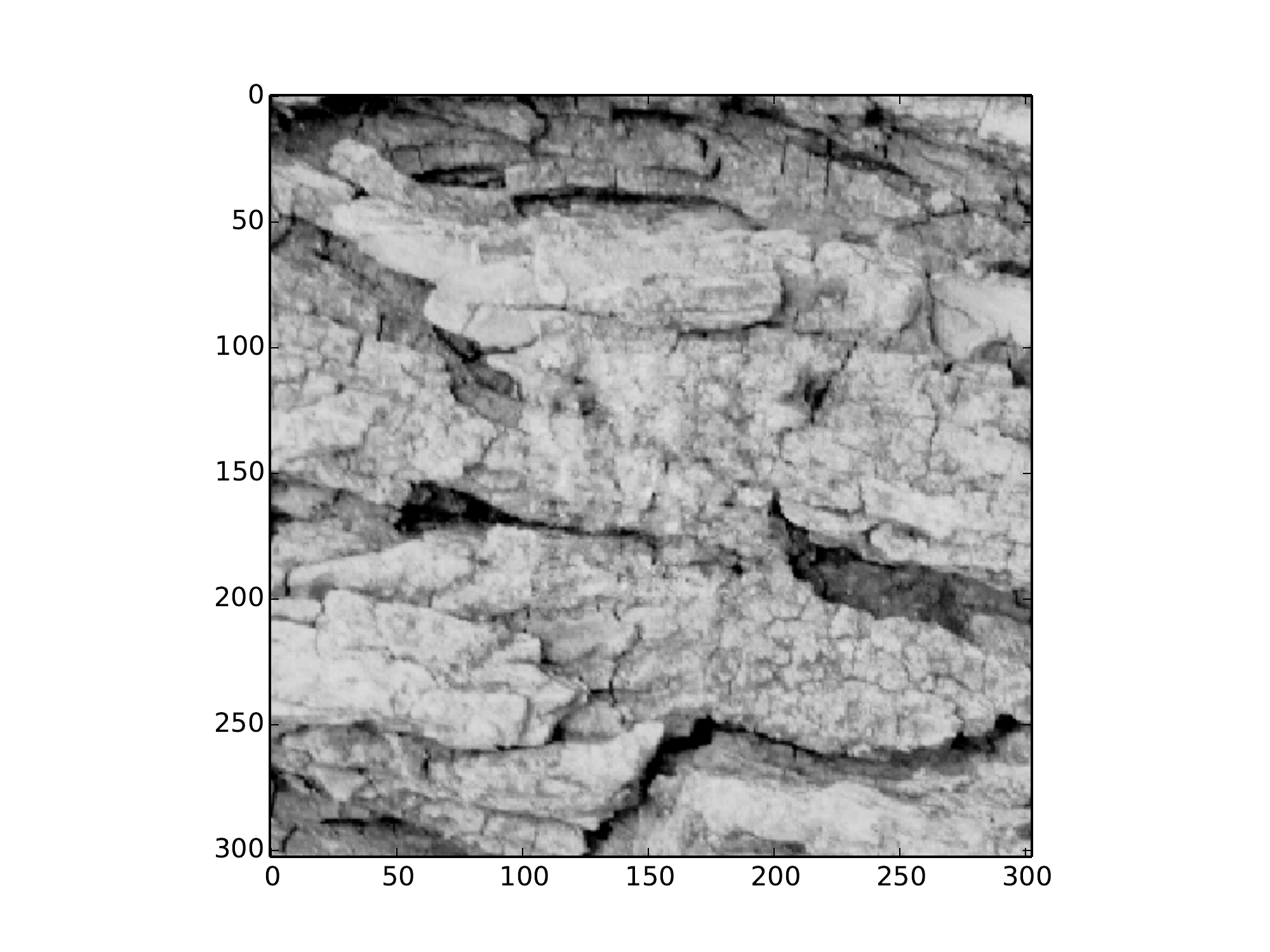}
\end{tabular}
\caption{
Inpainting.  {\em (a)} A bark image from \cite{lazebnik2005sparse}.  
{\em (b)} The same image with the central 100$\times$100 pixel region replaced with isotropic Gaussian noise.  This is the initialization $\tilde{p}\left( \mb x^{(T)} \right)$ for the reverse trajectory.  
{\em (c)} The central 100$\times$100 region has been inpainted using a diffusion probabilistic model trained on images of bark, by sampling from the posterior distribution over the missing region conditioned on the rest of the image.
Note the long-range spatial structure, for instance in the crack entering on the left side of the inpainted region. 
The sample from the posterior was generated as described in Section \ref{sec posterior}, where $r\left(\mb x^{(0)} \right)$ was set to a delta function for known data, and a constant for missing data.
}
\label{fig bark}
\end{figure*}
\section{Experiments}
\label{sec results}

We train diffusion probabilistic models on a variety of continuous datasets, and a binary dataset.
We then demonstrate sampling from the trained model and inpainting of missing data, and compare model performance against other techniques. 
In all cases the objective function and gradient were computed using Theano \cite{Bergstra2010}. Model training was with SFO \cite{sohl2014fast}, except for CIFAR-10. CIFAR-10 results used the open source implementation of the algorithm, and RMSprop for optimization.
The lower bound on the log likelihood provided by our model is reported for all datasets 
in Table \ref{tb K}.
A reference implementation of the algorithm utilizing Blocks \cite{Merrienboer2015} is available at \url{https://github.com/Sohl-Dickstein/Diffusion-Probabilistic-Models}.

\subsection{Toy Problems}

\subsubsection{Swiss Roll}

A diffusion  probabilistic model was built of a two dimensional swiss roll distribution, using a radial basis function network to generate
$\mb f_\mu\left( \mb x^{(t)}, t \right)$ and $\mb f_\Sigma\left( \mb x^{(t)}, t \right)$.
As illustrated in Figure \ref{fig swiss}, the swiss roll distribution was successfully learned. 
See Appendix Section \ref{sec swiss} for more details.

\subsubsection{Binary Heartbeat Distribution}

A diffusion  probabilistic model was trained on simple binary sequences of length 20, where a 1 occurs every 5th time bin, and the remainder of the bins are 0,
using a multi-layer perceptron to generate the Bernoulli rates $\mb f_b\left( \mb x^{(t)}, t \right)$ of the reverse trajectory.
The log likelihood under the true distribution is 
$\log_2\left( \frac{1}{5} \right) = -2.322$ bits per sequence.  As can be seen in Figure \ref{fig heartbeat} and Table \ref{tb K} learning was nearly perfect.
See Appendix Section \ref{sec heart} for more details.

\subsection{Images}\label{sec images}

We trained Gaussian diffusion probabilistic models on several image datasets. 
The multi-scale convolutional architecture shared by these experiments is described
in Appendix Section \ref{sec readout}, and illustrated in Figure \ref{fig architecture}.

\subsubsection{Datasets}

\paragraph{MNIST}

In order to allow a direct comparison against previous work on a simple dataset, we trained 
on MNIST digits \cite{MNIST}. 
Log likelihoods relative to \cite{Bengio2012,bengio2013deep,goodfellowgenerative} are given in Table \ref{tb ll compare}. 
Samples from the MNIST model are given in Appendix Figure \ref{fig mnist}. 
Our training algorithm provides an asymptotically consistent lower bound on the log likelihood. However most previous reported 
results on continuous MNIST log likelihood rely on Parzen-window based estimates computed from model samples. 
For this comparison we therefore estimate MNIST log likelihood using the Parzen-window code released with \cite{goodfellowgenerative}.

%

\paragraph{CIFAR-10}

A probabilistic model was fit to the training images for the CIFAR-10 challenge dataset \cite{Krizhevsky2009}. Samples from the trained model are provided in Figure \ref{fig cifar}.

\paragraph{Dead Leaf Images}

Dead leaf images \cite{Jeulin1997,Lee2001} consist of layered occluding circles, drawn from a power law distribution over scales. They have an analytically tractable structure, 
but capture many of the statistical complexities of natural images, and therefore provide a compelling test case for natural image models. 
As illustrated in Table \ref{tb ll compare} and Figure \ref{fig dead leaf}, we achieve state of the art performance on the dead leaves dataset.

\paragraph{Bark Texture Images}

A probabilistic model was trained on bark texture images (T01-T04) from \cite{lazebnik2005sparse}. For this dataset we demonstrate that 
it is straightforward to evaluate or generate from a posterior distribution, by inpainting a large region of missing data 
using a sample from the model posterior in Figure \ref{fig bark}.

\section{Conclusion}

We have introduced a novel algorithm for modeling probability distributions that enables exact sampling and evaluation of probabilities and demonstrated its effectiveness on a variety of toy and real datasets, including challenging natural image datasets. For each of these tests we used a similar basic algorithm, showing that our method can accurately model a wide variety of distributions. Most existing density estimation techniques must sacrifice modeling power in order to stay tractable and efficient, and sampling or evaluation are often extremely expensive. The core of our algorithm consists of estimating the reversal of a Markov diffusion chain which maps data to a noise distribution; as the number of steps is made large, the reversal distribution of each diffusion step becomes simple and easy to estimate. The result is an algorithm that can learn a fit to any data distribution, but which remains tractable to train, {\em exactly} sample from, and evaluate, and under which it is straightforward to manipulate conditional and posterior distributions.

\section*{Acknowledgements}
We thank Lucas Theis, Subhaneil Lahiri, Ben Poole, Diederik P. Kingma, Taco Cohen, Philip Bachman, and A\"{a}ron van den Oord for extremely helpful discussion, and Ian Goodfellow for Parzen-window code. We thank Khan Academy and the Office of Naval Research for funding Jascha Sohl-Dickstein, and we thank the Office of Naval Research and the Burroughs-Wellcome, Sloan, and James S. McDonnell foundations for funding Surya Ganguli.
\bibliography{icml2015_dpm}
\bibliographystyle{icml2015}

\onecolumn

\clearpage
\appendix

\normalsize

\part*{Appendix}

\setcounter{figure}{0} \renewcommand{\thefigure}{A.\arabic{figure}}
\setcounter{table}{0} \renewcommand{\thetable}{A.\arabic{table}}

\section{Conditional Entropy Bounds Derivation}\label{app entropy}

The conditional entropy $H_q\left( \mb X^{(t-1)} | \mb X^{(t)} \right)$ of a step in the reverse trajectory is
\begin{align}
H_q\left( \mb X^{(t-1)}, \mb X^{(t)} \right) &= H_q\left( \mb X^{(t)}, \mb X^{(t-1)} \right) \\
H_q\left( \mb X^{(t-1)} | \mb X^{(t)} \right) + H_q\left( \mb X^{(t)} \right) &= H_q\left( \mb X^{(t)} | \mb X^{(t-1)} \right) + H_q\left( \mb X^{(t-1)} \right) \\
H_q\left( \mb X^{(t-1)} | \mb X^{(t)} \right) &= H_q\left( \mb X^{(t)} | \mb X^{(t-1)} \right) + H_q\left( \mb X^{(t-1)} \right) - H_q\left( \mb X^{(t)} \right) \label{eq cond equal}
\end{align}

An upper bound on the entropy change can be constructed by observing that $\pi\left( \mb y \right)$ is the maximum entropy distribution.  This holds 
without qualification for the binomial distribution, and holds for variance 1 training data for the Gaussian case.  For the Gaussian case, training data 
must therefore be scaled to have unit norm for the following equalities to hold.  It need not be whitened.
The upper bound is derived as follows,
\begin{align}
H_q\left( \mb X^{(t)} \right) &\ge H_q\left( \mb X^{(t-1)} \right) \\
H_q\left( \mb X^{(t-1)} \right) - H_q\left( \mb X^{(t)} \right) &\le 0 \label{eq diff lb} \\
H_q\left( \mb X^{(t-1)} | \mb X^{(t)} \right) &\leq H_q\left( \mb X^{(t)} | \mb X^{(t-1)} \right)
.
\end{align}

A lower bound on the entropy difference can be established by observing that additional steps in a Markov chain do not increase the information 
available about the initial state in the chain, and thus do not decrease the conditional entropy of the initial state,
\begin{align}
\hspace{-2cm}
H_q\left( \mb X^{(0)} | \mb X^{(t)} \right) &\ge H_q\left( \mb X^{(0)} | \mb X^{(t-1)} \right) \\
\hspace{-2cm}
H_q\left( \mb X^{(t-1)} \right) - H_q\left( \mb X^{(t)} \right) &\ge
	H_q\left( \mb X^{(0)} | \mb X^{(t-1)} \right) + H_q\left( \mb X^{(t-1)} \right) - H_q\left( \mb X^{(0)} | \mb X^{(t)} \right) - H_q\left( \mb X^{(t)} \right) \\
\hspace{-2cm}
H_q\left( \mb X^{(t-1)} \right) - H_q\left( \mb X^{(t)} \right) &\ge
	H_q\left( \mb X^{(0)}, \mb X^{(t-1)} \right) - H_q\left( \mb X^{(0)}, \mb X^{(t)} \right)  \\
\hspace{-2cm}
H_q\left( \mb X^{(t-1)} \right) - H_q\left( \mb X^{(t)} \right) &\ge
	H_q\left( \mb X^{(t-1)} | \mb X^{(0)} \right)  - H_q\left( \mb X^{(t)} | \mb X^{(0)} \right) \label{eq diff ub} \\
\hspace{-2cm}
H_q\left( \mb X^{(t-1)} | \mb X^{(t)} \right) &\ge H_q\left( \mb X^{(t)} | \mb X^{(t-1)} \right) + H_q\left( \mb X^{(t-1)} | \mb X^{(0)} \right)  - H_q\left( \mb X^{(t)} | \mb X^{(0)} \right)
.
\end{align}

Combining these expressions, we bound the conditional entropy for a single step, 
\begin{align}
H_q\left( \mb X^{(t)} | \mb X^{(t-1)} \right)
&\ge
H_q\left( \mb X^{(t-1)} | \mb X^{(t)} \right)
\ge
H_q\left( \mb X^{(t)} | \mb X^{(t-1)} \right) + H_q\left( \mb X^{(t-1)} | \mb X^{(0)} \right)  - H_q\left( \mb X^{(t)} | \mb X^{(0)} \right)
,
\end{align}
where both the upper and lower bounds depend only on the conditional forward trajectory $\pcondtraj$, and can be analytically computed.

\section{Log Likelihood Lower Bound}\label{app bound}

The lower bound on the log likelihood is
\begin{align}
L
&\geq K \\
K &= \int d\mb x^{(0 \cdots T)} \ptraj  \log \left[ \qst \prod_{t=1}^T \frac{\qr}{\pf} \right] \\
\end{align}

\subsection{Entropy of $p\left( \mb X^{(T)} \right)$}

We can peel off the contribution from $p\left( \mb X^{(T)} \right)$, and rewrite it as an entropy,
\begin{align}
K&= \int d\mb x^{(0 \cdots T)} \ptraj  \sum_{t=1}^T\log \left[ \frac{\qr}{\pf} \right] 
	+ \int d\mb x^{(T)} \pmarg \log \qst \\
&= \int d\mb x^{(0 \cdots T)} \ptraj  \sum_{t=1}^T\log \left[ \frac{\qr}{\pf} \right] 
	+ \int d\mb x^{(T)} \pmarg \log \pi\left( \mb x^T \right) \\
.
\end{align}
By design, the cross entropy to $\pi\left( \mb x^{(t)}\right)$ is constant under our diffusion kernels, and equal to the entropy of $\qst$.  Therefore,
\begin{align}
K &=
  \sum_{t=1}^T \int d\mb x^{(0 \cdots T)} \ptraj  \log \left[ \frac{\qr}{\pf} \right] 
	- H_p\left( \mb X^{(T)} \right).
\end{align}

\subsection{Remove the edge effect at $t=0$}

In order to avoid edge effects, we set the final step of the reverse trajectory to be identical to the corresponding forward diffusion step,
\begin{align}
p\left( \mb x^{(0)} | \mb x^{(1)} \right) = q\left( \mb x^{(1)} | \mb x^{(0)} \right) \frac{ \pi\left( \mb x^{(0)}\right) }{ \pi\left( \mb x^{(1)}\right) } = T_\pi\left( \mb x^{(0)} | \mb x^{(1)}; \beta_1 \right).
\end{align}
We then use this equivalence to remove the contribution of the first time-step in the sum,
\begin{align}
K
&=  \sum_{t=2}^T \int d\mb x^{(0 \cdots T)} \ptraj  \log \left[ \frac{\qr}{\pf} \right] 
	+ \int d\mb x^{(0)} d\mb x^{(1)} q\left( \mb x^{(0)}, \mb x^{(1)} \right)  \log \left[ \frac{
		q\left( \mb x^{(1)} | \mb x^{(0)} \right) \pi\left( \mb x^{(0)}\right)
		}{
		q\left( \mb x^{(1)} | \mb x^{(0)} \right) \pi\left( \mb x^{(1)}\right)
		} \right]
	- H_p\left( \mb X^{(T)} \right)
	\\
&=  \sum_{t=2}^T \int d\mb x^{(0 \cdots T)} \ptraj  \log \left[ \frac{\qr}{\pf} \right] 
	- H_p\left( \mb X^{(T)} \right)
	,
\end{align}
where we again used the fact that by design $-\int d\mb x^{(t)} q\left( \mb x^{(t)} \right) \log \pi\left( \mb x^{(t)}\right) = H_p\left( \mb X^{(T)} \right)$ is a constant for all $t$.

\subsection{Rewrite in terms of posterior $q\left( \mb x^{(t-1)} | \mb x^{(0)} \right)$}

Because the forward trajectory is a Markov process,
\begin{align}
K &= \sum_{t=2}^T \int d\mb x^{(0 \cdots T)} \ptraj  \log \left[ \frac{\qr}{q\left( \mb x^{(t)} | \mb x^{(t-1)}, \mb x^{(0)} \right)} \right] 
	- H_p\left( \mb X^{(T)} \right).
\end{align}
Using Bayes' rule we can rewrite this in terms of a posterior and marginals from the forward trajectory,
\begin{align}
K
&= \sum_{t=2}^T \int d\mb x^{(0 \cdots T)} \ptraj  \log \left[ 
		\frac{\qr}{q\left( \mb x^{(t-1)} | \mb x^{(t)}, \mb x^{(0)} \right)} \frac{q\left( \mb x^{(t-1)} | \mb x^{(0)} \right)}{q\left( \mb x^{(t)} | \mb x^{(0)} \right)}
	\right] 
	- H_p\left( \mb X^{(T)} \right).
\end{align}

\subsection{Rewrite in terms of KL divergences and entropies}

We then recognize that several terms are conditional entropies,
\begin{align}
K
&= \sum_{t=2}^T \int d\mb x^{(0 \cdots T)} \ptraj  \log \left[ 
		\frac{\qr}{q\left( \mb x^{(t-1)} | \mb x^{(t)}, \mb x^{(0)} \right)} 
	\right] 
	+ \sum_{t=2}^T \left[ H_q\left( \mb X^{(t)} | \mb X^{(0)} \right) - H_q\left( \mb X^{(t-1)} | \mb X^{(0)} \right)\right]
	- H_p\left( \mb X^{(T)} \right) \\
&= \sum_{t=2}^T \int d\mb x^{(0 \cdots T)} \ptraj  \log \left[ 
		\frac{\qr}{q\left( \mb x^{(t-1)} | \mb x^{(t)}, \mb x^{(0)} \right)} 
	\right] 
	+ H_q\left( \mb X^{(T)} | \mb X^{(0)} \right) - H_q\left( \mb X^{(1)} | \mb X^{(0)} \right)
	- H_p\left( \mb X^{(T)} \right)
.
\end{align}
Finally we transform the log ratio of probability distributions into a KL divergence,
\begin{align}
K
&= -\sum_{t=2}^T \int d\mb x^{(0)}d\mb x^{(t)} q\left( \mb x^{(0)}, \mb x^{(t)} \right)
D_{KL}\left( 
		q\left( \mb x^{(t-1)} | \mb x^{(t)}, \mb x^{(0)} \right)
			\middle|\middle|
		\qr
	\right)  \\ \nonumber & \qquad
	+ H_q\left( \mb X^{(T)} | \mb X^{(0)} \right) - H_q\left( \mb X^{(1)} | \mb X^{(0)} \right)
	- H_p\left( \mb X^{(T)} \right)
.
\end{align}
Note that the entropies can be analytically computed, and the KL divergence can be analytically computed given $\mb x^{(0)}$ and $\mb x^{(t)}$.

\setcounter{figure}{0} \renewcommand{\thefigure}{App.\arabic{figure}}
\setcounter{table}{0} \renewcommand{\thetable}{App.\arabic{table}}
\begin{figure}
\centering
\includegraphics[width=0.8\linewidth]{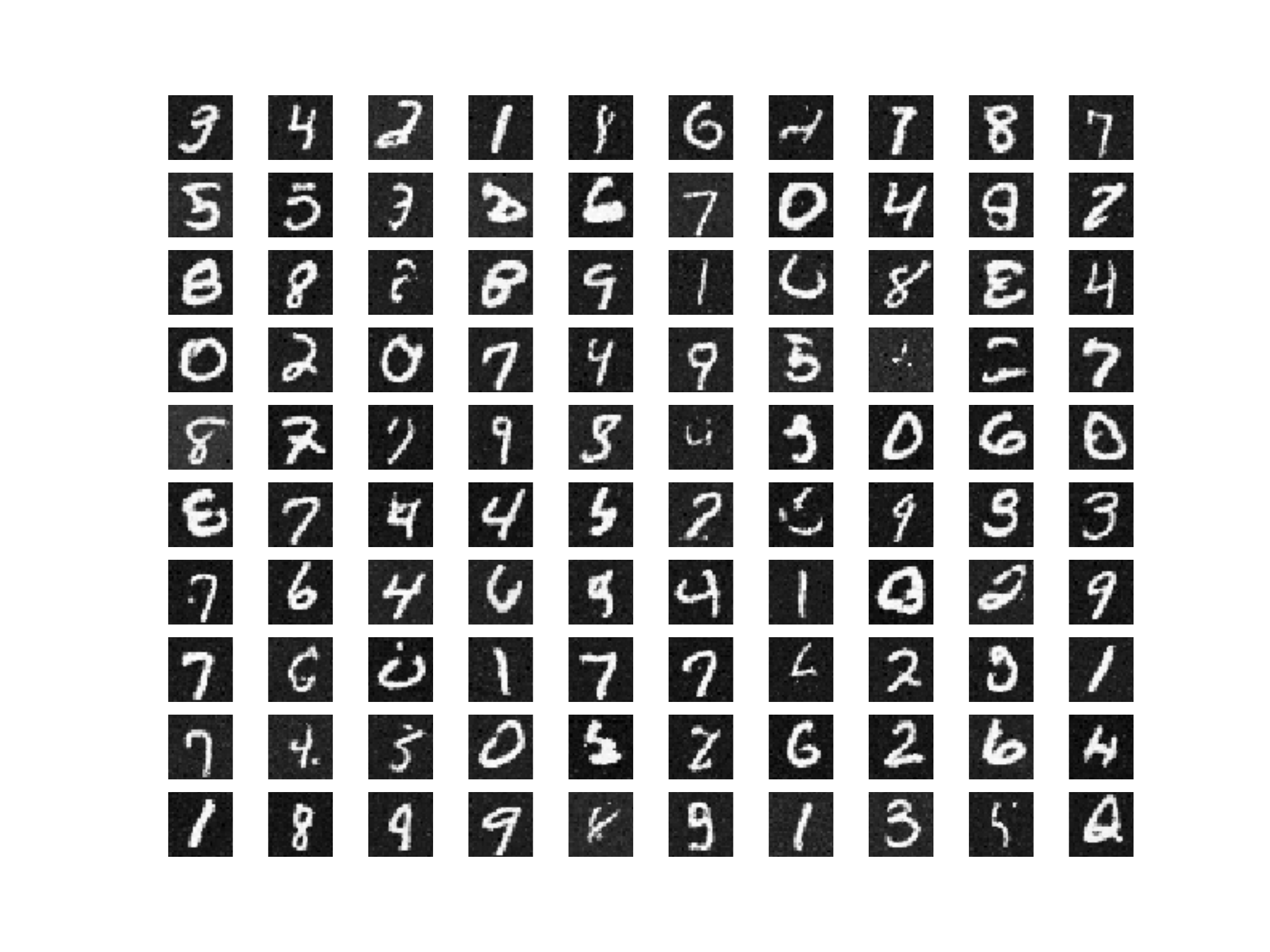} 
\caption{
Samples from a diffusion probabilistic model trained on MNIST digits. 
Note that unlike many MNIST sample figures, these are true samples rather than the mean of the Gaussian or binomial distribution from which samples would be drawn.
}
\label{fig mnist}
\end{figure}

\begin{landscape}
\thispagestyle{empty}
\begin{table*}
\renewcommand{\arraystretch}{1.45}
\hspace{-.8in}
\begin{tabular}{p{0.22\linewidth}|r|l|l}
 			& & \em Gaussian & \em Binomial \\ \hline
Well behaved (analytically tractable) distribution & $\ptarget = $ & $\mc N \left(\mb x^{(T)} ;  \mb 0, \mb I \right)$  & $\mc B\left( \mb x^{(T)} ; 0.5 \right)$  \\
Forward diffusion kernel & $\pf = $ & $\mc N \left( \mb x^{(t)} ; \mb x^{(t-1)} \sqrt{1 - \beta_t}, \mb I \beta_t \right)$  & $\mc B \left( \mb x^{(t)} ; \mb x^{(t-1)} \left(1 - \beta_t\right) + 0.5 \beta_t \right)$ \\
Reverse diffusion kernel & $\qr =$ & $\mc N \left( \mb x^{(t-1)} ;\mb f_\mu\left( \mb x^{(t)}, t \right), \mb f_\Sigma\left( \mb x^{(t)}, t \right) \right)$  & $\mc B \left( \mb x^{(t-1)} ; \mb f_b\left( \mb x^{(t)}, t \right) \right)$  \\
Training targets & $$ & $\mb f_\mu\left( \mb x^{(t)}, t \right)$, $\mb f_\Sigma\left( \mb x^{(t)}, t \right)$, $\beta_{1\cdots T}$   & $\mb f_b\left( \mb x^{(t)}, t \right)$  \\
Forward distribution & $\ptraj = $ & \multicolumn{2}{c}{$\pst \prod_{t=1}^T \pf$}  \\
Reverse distribution & $\qtraj = $ & \multicolumn{2}{c}{$\ptarget \prod_{t=1}^T \qr$}  \\
Log likelihood & $L =$ & \multicolumn{2}{c}{$\int d\mb x^{(0)} \pst \log \qmarg$} \\
Lower bound on log likelihood & $K =$ & \multicolumn{2}{c}{
	$-\sum_{t=2}^T \mathbb E_{q\left( \mb x^{(0)}, \mb x^{(t)} \right)}\left[
	D_{KL}\left( 
		q\left( \mb x^{(t-1)} | \mb x^{(t)}, \mb x^{(0)} \right)
			\middle|\middle|
		\qr
	\right)
	\right]
	+ H_q\left( \mb X^{(T)} | \mb X^{(0)} \right) - H_q\left( \mb X^{(1)} | \mb X^{(0)} \right)
	- H_p\left( \mb X^{(T)} \right) \nonumber
	$
	}  \\
Perturbed reverse diffusion kernel & $\qrtil =$ & 
	$	\mc N \left( \mb \mb x^{(t-1)} ;
		\mb f_\mu\left( \mb x^{(t)}, t \right) + \mb f_\Sigma\left( \mb x^{(t)}, t \right)
		\pd{
			\log r\left( \mb x^{(t-1)'} \right)
			}{
			\mb x^{(t-1)'}
			}\bigg|_{\mb x^{(t-1)'}=f_\mu\left( \mb x^{(t)}, t \right)}
			,
		\mb f_\Sigma\left( \mb x^{(t)}, t \right) 
		\right)$
	 & 
	 $\mc B \left( x_i^{(t-1)} ; 
	 	\frac{
			 c_i^{t-1} d_i^{t-1}
		}{
			x_i^{t-1} d_i^{t-1} + (1-c_i^{t-1})(1-d_i^{t-1})
		}
	 \right)$  
\end{tabular}
\caption{
The key equations in this paper for the specific cases of Gaussian and binomial diffusion processes.  $\mc N\left( u; \mu, \Sigma \right)$ is 
a Gaussian distribution with mean $\mu$ and covariance $\Sigma$.  $\mc B\left( u; r \right)$ is the distribution for a single Bernoulli trial, with $u=1$ 
occurring with probability $r$, and $u=0$ occurring with probability $1-r$. Finally, for the perturbed Bernoulli trials
 $b_i^{t} = \mb x^{(t-1)} \left(1 - \beta_t\right) + 0.5 \beta_t$,
 $c_i^{t} = \left[\mb f_b\left( \mb x^{(t+1)}, t \right)\right]_i$,
 and $d_i^t = r\left( x_i^{(t)} = 1\right)$, and the distribution is given for a single bit $i$.
\label{tab diff}
}
\end{table*}
\end{landscape}

\setcounter{figure}{0} \renewcommand{\thefigure}{C.\arabic{figure}}
\setcounter{table}{0} \renewcommand{\thetable}{C.\arabic{table}}

\section{Perturbed Gaussian Transition}
\label{sec perturb derive}

We wish to compute $\tilde{p}\left( \mb x^{(t-1)} \mid \mb x^{(t)} \right)$. For notational simplicity, let $\mu = \mb f_\mu\left( \mb x^{(t)}, t \right)$, $\Sigma = \mb f_\Sigma\left( \mb x^{(t)}, t \right)$, and $\mb y = \mb x^{(t-1)}$. Using this notation,
\begin{align}
\tilde{p}\left( \mb y \mid \mb x^{(t)} \right)
&\propto 
	p\left( \mb y \mid \mb x^{(t)} \right)
	r\left( \mb y \right) \\
&=	\mc N \left( \mb y ;\mu, \Sigma \right)
	r\left( \mb y \right)
.
\end{align}
We can rewrite this in terms of energy functions, where $E_r\left( \mb y \right) = -\log r\left( \mb y \right)$,
\begin{align}
\tilde{p}\left( \mb y \mid \mb x^{(t)} \right)
&\propto
	\exp\left[
		-E\left( \mb y \right) 
	\right] \\
E\left( \mb y \right) &=\frac{1}{2} \left( \mb y - \mu \right)^T \Sigma^{-1} \left( \mb y - \mu \right) +  E_r\left( \mb y \right)
.
\end{align}

If $E_r\left( \mb y \right)$ is smooth relative to $\frac{1}{2} \left( \mb y - \mu \right)^T \Sigma^{-1} \left( \mb y - \mu \right)$, then we can approximate it using its Taylor expansion around $\mu$. One sufficient condition is that the eigenvalues of the Hessian of $E_r\left( \mb y \right)$ are everywhere much smaller magnitude than the eigenvalues of $\Sigma^{-1}$. We then have
\begin{align}
E_r\left( \mb y \right) & \approx E_r\left( \mu \right) + \left( \mb y - \mu \right) \mb g
\end{align}
where $\mb g = \pd{E_r\left( \mb y' \right)}{\mb y'} \bigg|_{\mb y'=\mu}$. Plugging this in to the full energy,
\begin{align}
E\left( \mb y \right) &\approx \frac{1}{2} \left( \mb y - \mu \right)^T \Sigma^{-1} \left( \mb y - \mu \right) + \left( \mb y - \mu \right)^T \mb g + \text{constant} \\
&= \frac{1}{2} \mb y^T \Sigma^{-1} \mb y 
	- \frac{1}{2} \mb y^T \Sigma^{-1} \mu
	- \frac{1}{2} \mu^T \Sigma^{-1} \mb y
	+ \frac{1}{2} \mb y^T \Sigma^{-1} \Sigma \mb g
	+ \frac{1}{2} \mb g^T \Sigma  \Sigma^{-1} \mb y
	+ \text{constant}
	\\
&=	\frac{1}{2}
	\left( \mb y - \mu + \Sigma \mb g\right)^T 
	\Sigma^{-1} 
	\left( \mb y - \mu + \Sigma \mb g \right)
	+ \text{constant}
.
\end{align}
This corresponds to a Gaussian,
\begin{align}
\tilde{p}\left( \mb y \mid \mb x^{(t)} \right)
&\approx 
	\mc N \left( \mb y ;
		\mu - \Sigma \mb g,
		\Sigma 
		\right)
.
\end{align}
Substituting back in the original formalism, this is,
\begin{align}
\tilde{p}\left( \mb x^{(t-1)} \mid \mb x^{(t)} \right)
&\approx 
	\mc N \left( \mb \mb x^{(t-1)} ;
		\mb f_\mu\left( \mb x^{(t)}, t \right) + \mb f_\Sigma\left( \mb x^{(t)}, t \right)
		\pd{
			\log r\left( \mb x^{(t-1)'} \right)
			}{
			\mb x^{(t-1)'}
			}\Bigg|_{\mb x^{(t-1)'}=f_\mu\left( \mb x^{(t)}, t \right)}
			,
		\mb f_\Sigma\left( \mb x^{(t)}, t \right) 
		\right)
.
\end{align}

\twocolumn

\setcounter{figure}{0} \renewcommand{\thefigure}{D.\arabic{figure}}
\setcounter{table}{0} \renewcommand{\thetable}{D.\arabic{table}}

\section{Experimental Details}\label{app experiments}

\subsection{Toy Problems}

\subsubsection{Swiss Roll}
\label{sec swiss}
A probabilistic model was built of a two dimensional swiss roll distribution. 
The generative model $\qtraj$ consisted of 40 time steps of Gaussian 
diffusion initialized at an identity-covariance Gaussian distribution.  
A (normalized) radial basis function network with a single hidden layer and 16 hidden units was trained to generate the mean and covariance 
functions $\mb f_\mu\left( \mb x^{(t)}, t \right)$ and a diagonal $\mb f_\Sigma\left( \mb x^{(t)}, t \right)$ for the reverse trajectory. 
The top, readout, layer for each function was 
learned independently for each time step, but for all other layers weights were shared across all time steps and both functions. 
The top layer output of $\mb f_\Sigma\left( \mb x^{(t)}, t \right)$ was passed through a sigmoid to restrict it between 0 and 1. 
As can be seen in Figure \ref{fig swiss}, the swiss roll distribution was successfully learned.

\subsubsection{Binary Heartbeat Distribution}
\label{sec heart}
A probabilistic model was trained on simple binary sequences of length 20, where a 1 occurs every 5th time bin, and the remainder of the bins are 0.  
The generative model consisted of 2000 time steps of binomial diffusion initialized at an independent binomial distribution with the 
same mean activity as the data ($p\left( x_i^{(T)} = 1 \right) = 0.2$). 
A multilayer perceptron with sigmoid nonlinearities, 20 input units 
and three hidden layers with 
50 units each was trained to generate the Bernoulli rates $\mb f_b\left( \mb x^{(t)}, t \right)$ of the reverse trajectory.  
The top, readout, layer was 
learned independently for each time step, but for all other layers weights were shared across all time steps. 
The top layer output was passed through a sigmoid to restrict it between 0 and 1. 
As can be seen in 
Figure \ref{fig heartbeat}, the heartbeat distribution was successfully learned.  The log likelihood under the true generating process is 
$\log_2\left( \frac{1}{5} \right) = -2.322$ bits per sequence.  As can be seen in Figure \ref{fig heartbeat} and Table \ref{tb K} learning was nearly perfect.

\subsection{Images}

\subsubsection{Architecture}\label{sec readout}
\paragraph{Readout}\label{sec readout}

In all cases, a convolutional network was used to produce a vector of outputs $\mb y_i \in \mc R^{2J}$ for each image pixel $i$. 
The entries in $\mb y_i$ are divided into two equal sized subsets, $\mb y^\mu$ and $\mb y^\Sigma$. 

\paragraph{Temporal Dependence}

The convolution output $\mb y^\mu$ is used as per-pixel weighting coefficients in a sum over time-dependent ``bump'' functions, 
generating an output $\mb z^\mu_i \in \mc R$ for each pixel $i$,
\begin{align}
\mb z^\mu_i &= \sum_{j=1}^J \mb y^\mu_{ij} g_j\left(t\right)
.
\end{align}
The bump functions consist of
\begin{align}
g_j\left(t\right) &= \frac{
	\exp\left( 
		-\frac{1}{2 w^2} \left( t - \tau_j \right)^2
	\right)
}{
	\sum_{k=1}^J
	\exp\left( 
		-\frac{1}{2 w^2} \left( t - \tau_k \right)^2
	\right)
}
,
\end{align}
where $\tau_j \in (0, T)$ is the bump center, and $w$ is the spacing between bump centers. 
$\mb z^\Sigma$ is generated in an identical way, but using $y^\Sigma$.

For all image experiments a number of timesteps $T=1000$ was used, except for the bark dataset which used $T=500$.

\paragraph{Mean and Variance}

Finally, these outputs are combined to produce a diffusion mean and variance prediction for each pixel $i$, 
\begin{align}
\Sigma_{ii} &= \sigma\left( z^\Sigma_i + \sigma^{-1}\left( \beta_t \right) \right), \\
\mu_i &= \left( x_i - z^\mu_i \right)\left(1 - \Sigma_{ii}\right) + z^\mu_i
.
\end{align}
where both $\Sigma$ and $\mu$ are parameterized as a perturbation around the forward diffusion kernel $T_\pi\left( \mb x^{(t)} | \mb x^{(t-1)}; \beta_t \right)$, 
and $z^\mu_i$ is the mean of the equilibrium distribution that would result from applying $\qr$ many times. $\Sigma$ is restricted to be a diagonal matrix.

\paragraph{Multi-Scale Convolution}\label{sec multiscale}

We wish to accomplish goals that are often achieved with pooling networks -- specifically, we wish to discover and make use of long-range and multi-scale dependencies 
in the training data. However, since the network output is a vector of coefficients for every pixel 
it is important to generate a full resolution rather than down-sampled feature map. We therefore define multi-scale-convolution layers that consist of the following steps:
\begin{enumerate}\itemsep1pt \parskip0pt \parsep0pt
  \item Perform mean pooling to downsample the image to multiple scales. Downsampling is performed in powers of two.
  \item Performing convolution at each scale.
  \item Upsample all scales to full resolution, and sum the resulting images.
  \item Perform a pointwise nonlinear transformation, consisting of a soft relu ($\log\left[ 1 + \exp\left(\cdot \right)\right]$). 
\end{enumerate}
The composition of the first three linear operations resembles convolution by a multiscale convolution kernel, up to blocking artifacts introduced by upsampling. This method of achieving multiscale convolution was described in \cite{Barron2013}.

\paragraph{Dense Layers}

Dense (acting on the full image vector) and kernel-width-1 convolutional (acting separately on the feature vector for each pixel) layers share the same form. 
They consist of a linear transformation, followed by a tanh nonlinearity.

\begin{figure}
\centering
\adjincludegraphics[width=\linewidth]{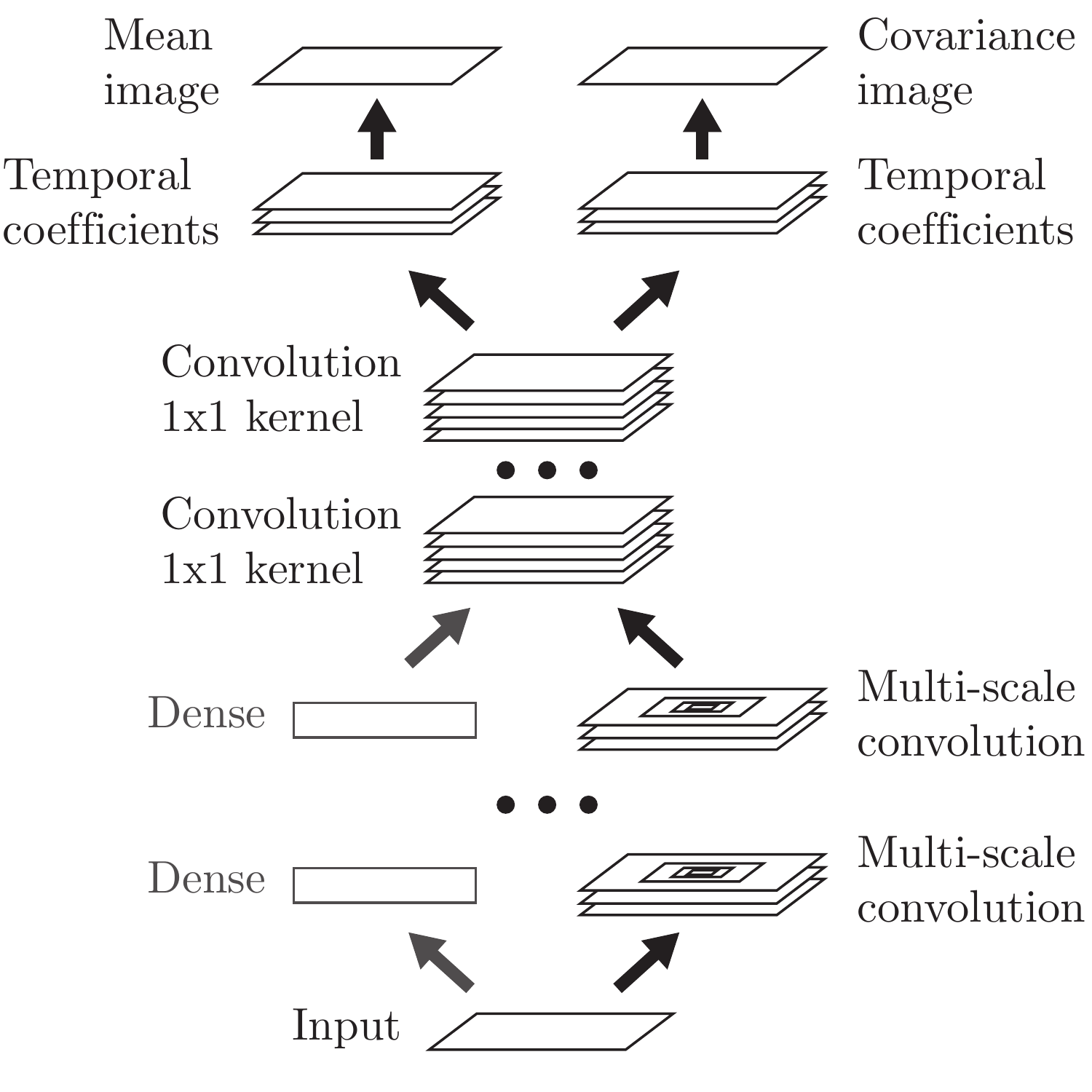}
\caption{
Network architecture for mean function $\mb f_\mu\left( \mb x^{(t)}, t \right)$ and covariance function $\mb f_\Sigma\left( \mb x^{(t)}, t \right)$, for experiments 
in Section \ref{sec images}. The input image $\mb x^{(t)}$ passes through several layers of multi-scale convolution (Section \ref{sec multiscale}). 
It then passes through several convolutional layers with $1\times 1$ kernels. This is equivalent to a dense transformation performed on each pixel. 
A linear transformation generates coefficients for readout of both mean $\mu^{(t)}$ and covariance $\Sigma^{(t)}$ for each pixel. 
Finally, a time dependent readout function converts those coefficients into mean and covariance images, as described in Section \ref{sec readout}. 
For CIFAR-10 a dense (or fully connected) pathway was used in parallel to the multi-scale convolutional pathway. For MNIST, the dense pathway was 
used to the exclusion of the multi-scale convolutional pathway.
}
\label{fig architecture}
\end{figure}

\end{document}